\documentclass[journal, 10pt, comsoc]{IEEEtran}
\usepackage{amsmath, amsfonts, algorithmic, algorithm, array, fancyhdr}
\usepackage[caption=false,font=small,labelfont=sf,textfont=sf]{subfig}
\usepackage{booktabs}
\usepackage{textcomp}
\usepackage{stfloats}
\usepackage{url}
\usepackage{verbatim}
\usepackage{graphicx}
\usepackage{cite}
\usepackage{color}
\newtheorem{myOpt}{\textbf{Problem}}

\begin{document}

\title{Deep Learning-Empowered Semantic Communication Systems with a Shared Knowledge Base}

\author{\IEEEauthorblockN{Peng Yi, Yang Cao, Xin Kang, \emph{Senior Member, IEEE}, and Ying-Chang Liang, \emph{Fellow, IEEE}} \\ 
\thanks{
    P. Yi, and X. Kang are with the National Key Laboratory of Wireless Communications, and the Center for Intelligent Networking and Communications (CINC), University of Electronic Science and Technology of China (UESTC), Chengdu 611731, China. (email: yipengcd@outlook.com, kangxin83@gmail.com).}
\thanks{
    Y. Cao is with the School of Information Science and Technology, Southwest Jiaotong University, Chengdu 611756, China. (e-mail: cyang9502@gmail.com).}
\thanks{
    Y.-C. Liang is with the Center for Intelligent Networking and Communications (CINC), University of Electronic Science and Technology of China (UESTC), Chengdu 611731, China. (email:liangyc@ieee.org).}
}

\maketitle
\thispagestyle{fancy}
\lhead{\small \copyright 10.1109/TWC.2023.3330744 IEEE. Personal use of this material is permitted. Permission from IEEE must be obtained for all other uses, in any current or future media, including reprinting/republishing this material for advertising or promotional purposes, creating new collective works, for resale or redistribution to servers or lists, or reuse of any copyrighted component of this work in other works.}
\renewcommand{\headrulewidth}{0mm}

\begin{abstract}
Deep learning-empowered semantic communication is regarded as a promising candidate for future 6G networks. Although existing semantic communication systems have achieved superior performance compared to traditional methods, the end-to-end architecture adopted by most semantic communication systems is regarded as a black box, leading to the lack of explainability. To tackle this issue, in this paper, a novel semantic communication system with a shared knowledge base is proposed for text transmissions. Specifically, a textual knowledge base constructed by inherently readable sentences is introduced into our system. With the aid of the shared knowledge base, the proposed system integrates the message and corresponding knowledge from the shared knowledge base to obtain the residual information, which enables the system to transmit fewer symbols without semantic performance degradation. In order to make the proposed system more reliable, the semantic self-information and the source entropy are mathematically defined based on the knowledge base. Furthermore, the knowledge base construction algorithm is developed based on a similarity-comparison method, in which a pre-configured threshold can be leveraged to control the size of the knowledge base. Moreover, the simulation results have demonstrated that the proposed approach outperforms existing baseline methods in terms of transmitted data size and sentence similarity.
\end{abstract}


\section{Introduction} 

\IEEEPARstart{S}{ince} Shannon established information theory in 1949 \cite{DBLP:journals/bstj/Shannon48}, researchers have been dedicated to designing coding schemes based on the Shannon-Weaver model of communication, in which semantic aspects of communication are considered irrelevant to the engineering implementations of communication systems. With the development of communication technologies, various source coding schemes have approached the source entropy or rate-distortion function, and advanced channel coding schemes such as low-density parity-check (LDPC) \cite{DBLP:journals/comsur/FangBGL15} and polar codes \cite{DBLP:journals/comsur/EgilmezXMH20} have approached Shannon limited. Further compressing the source at the bit level is therefore costly and ineffective \cite{DBLP:journals/cm/KaiNiu22}. 

Nevertheless, with the advent of the sixth generation (6G) \cite{SIX-Trust_KangXin}, the existing coding methods face stringent challenges when dealing with increasing system capacity and expanding coverage. To be specific, 6G is expected to provide services to massive devices (more than 10$\mathrm{million/km}^2$) and to be reliable for ultra-high area traffic density (more than 100$\mathrm{Tb/s/km}^2$) \cite{DBLP:journals/wc/ChenLSKCP20}, which cannot be cost-effectively satisfied by conventional technology enhancements due to spectrum resources and power constraints. On the other hand, to support ubiquitous global connections, non-terrestrial networks (NTNs) are recognized as a critical component of 6G \cite{DBLP:journals/network/GiordaniZ21}. In such a case, wireless communications between low-earth orbit (LEO) satellites and terrestrial devices suffer from extremely high path loss, leading to an extremely low signal-to-noise ratio (SNR) signal received by ground users. 

These issues motivate a paradigm shift from conventional communications to semantic communications \cite{zhang2022toward, wheeler2022engineering}, which extract semantic meanings behind bits for transmission and recover the semantic information at the receiver through minimizing the semantic error. There are two main advantages of semantic communications over conventional communications. One is that more efficient compression of the source can be achieved through transmitting only semantic information with few bits \cite{DBLP:conf/globecom/HuangTGL21}. Secondly, although syntactic errors may exist in a semantic communication system due to the unavoidable channel noise, semantic meanings may not change, which allows the system to achieve superior performance in the low SNR regime \cite{DBLP:journals/wc/LuoCG22}. However, the investigation of semantic communications has experienced decades of stagnation due to the absence of a mathematical model for semantic information theory in engineering \cite{DBLP:journals/corr/abs-2201-01389}. 

Recently, deep learning (DL) \cite{DBLP:books/aw/RN2020} has empowered devices with cognitive abilities and make it possible to realize semantic communications at the engineering level. On the one hand, by regarding communication system design as an end-to-end reconstruction task, isolated functions (e.g., source/channel coding and modulation) are replaced with deep neural networks (DNNs), which allows the transmitter and receiver to be jointly optimized for information compression and noise resistance \cite{DBLP:journals/tccn/OSheaH17}. On the other hand, notable success in natural language processing (NLP) and computer vision (CV) fields demonstrates the ability of DL to extract semantic meanings behind texts and images. Therefore, with the advantages of DL in physical layer modeling and semantic feature extraction, semantic communications have received much attention in realizing intelligent wireless communications.

\subsection{Related Work}
Without loss of generality, most of the existing semantic systems can be divided into two categories according to different communication modalities, i.e., human-to-human (H2H) and human-to-machine (H2M) communications \cite{DBLP:journals/jcin/LanWZZCPH21, DBLP:journals/cn/StrinatiB21}. 
To be specific, H2H communications aim to deliver meanings accurately over a channel for message exchanges between two human beings \cite{wheeler2022engineering}. While in H2M communications, the message sent by a human being should be correctly interpreted by a machine toward performing specific tasks such as speech to text transmission \cite{DBLP:conf/icc/HanY0H022}, image classification \cite{DBLP:journals/tcom/KangSGQY22}, spectrum sensing \cite{DBLP:conf/mlsp/YiCXL22} and visual question answering \cite{DBLP:journals/jsac/XieQTL22}. In other words, the former keeps the modality of the message unchanged (e.g., from image to image), while the latter changes the message modality (e.g., from image to classifications). 
Although, H2M semantic communication systems work well for specific tasks, it is worth pointing out that the DNNs of the transceiver is designed based on the objective of task and should be reconfigured in different scenarios. In this case, multiple transmitters or multiple receivers are needed to accomplish difference tasks \cite{DBLP:journals/icl/HuZZWHZ22}. Therefore, compared to H2M semantic communications, H2H ones are more generalized in terms of tasks, since the received information can be reused for various downstream tasks \cite{DBLP:journals/jcin/LanWZZCPH21}. 

Among H2H schemes, a DL-based end-to-end semantic communication framework, namely DeepSC, was first proposed for text transmission in \cite{DBLP:journals/tsp/XieQLJ21} as the landmark of the revolution of semantic communications. Besides, sentence similarity was adopted as a new metric to justify the performance of semantic communication systems accurately. Designed to maximize the system capacity while simultaneously minimizing semantic errors, DeepSC outperformed traditional methods with respect to sentence similarity metrics, especially in the low SNR regime. To further validate the generality of the DeepSC framework, DeepSC-S \cite{DBLP:journals/jsac/WengQ21}, as an extended version of DeepSC, was proposed for speech signals transmission, in which the attention mechanism was leveraged to identify the essential speech information. 
Based on the DeepSC structure, to further improve the performance of semantic communication systems and mitigate the effects of channel noise, many techniques have been introduced at the transmitter and receiver ends, respectively. 
To be specific, at the transmitter side, the channel state information \cite{DBLP:journals/jsac/XieQ21} and the semantic importance distribution \cite{DBLP:journals/jsac/WangCLSNPC22} were investigated to tackle varying channel conditions and perform resource allocation, respectively. At the receiver side, undesirable distortions were alleviated through introducing a hybrid automatic repeat request mechanism \cite{DBLP:journals/tcom/JiangWJL22}, a semantic distortion correction module \cite{DBLP:journals/corr/abs-2112-03093} or an iterative decoding architecture \cite{yao2022semantic}. Through the above techniques, the maximum preservation of semantic information can be provided by H2H semantic communications even at a low SNR or a high compression rate.

Although the aforementioned H2H communication schemes achieve better performance compared to the traditional methods, several issues remain unsolved. Firstly, due to the fact that existing semantic communication systems are based on end-to-end architecture, the background knowledge is stored as millions of parameters in the DNNs, and thus the semantic information is searched from an extremely high-dimensional semantic space. Additionally, the whole DNNs need to be retrained frequently when the DNN-based background knowledge is not suitable for various deployment scenarios, leading to a time-consuming process. Luckily, an interpretable knowledge base \cite{ICC2023} can guide users to extract the semantic information from a squeezed semantic space, providing ``explainability'' benefits that make the public more willing to trust the model \cite{ribeiro2016should}. 
Secondly, there is a lack of semantic information theory to support technical implementation of semantic communication theory. To be specific, the widely studied semantic information theories based on logical probability \cite{bao2011towards} and fuzzy sets \cite{DBLP:journals/iandc/LucaT74} are hard to be implemented at the technical level, since precise logical reasoning mechanisms and complex world models are required. On the other hand, DL-based semantic communication systems directly learn from the collected data and build encoders and decoders, in which the coding and decoding operations, and corresponding gains cannot be theoretically illustrated by the current semantic information theory. Toward designing a DL-enabled semantic communication systems, it is in urgent need to render the coding and decoding operations more interpretable and explainable in terms of semantic information theory, which could also provide a guideline for our future research.

\subsection{Main Contributions}
To overcome these limitations, in this paper, a novel semantic communication system with a shared knowledge base is proposed for text transmission. To be specific, instead of mapping the message directly into the latent space, the transmitter in our system integrates the message with the corresponding knowledge from the shared knowledge base to obtain the residual information that needs to be transmitted. Then, at the receiver side, the received residual information is fused with the same knowledge to recover the message. The main contributions of this work are as follows.
\begin{enumerate}{}{}
\item {In this paper, a novel semantic communication framework is proposed for text transmission over a physical channel. Different from existing semantic communication systems, our system consists of three participants, i.e., transmitter, receiver, and a shared knowledge base. Since the shared knowledge base is constructed by inherently readable sentences, our proposed system is more explainable than most existing semantic communication systems based on the end-to-end architecture, which are normally regarded as black-box systems.}
\item {Considering that most existing semantic communication systems lack information-theoretic support, a mathematical interpretation of the proposed framework is given, making our system more reliable. Particularly, we observe that it is not reasonable to define semantic entropy in terms of the probability of occurrence of semantic information. With the aid of the knowledge base, the self-information of the message thus can be defined by the probability of occurrence of its most similar knowledge and the distance between them.}
\item {An effective algorithm to develop a textual knowledge base is proposed based on the sentence similarity metric, in which a pre-configured threshold $\theta$ is leveraged to control the size of the knowledge base and the amount of residual information.}
\item {The semantic communication system with a shared knowledge base is implemented for text transmission based on the self-attention mechanism. Numerical simulation results have demonstrated that the proposed system outperforms other benchmarks in terms of transmitted data size and sentence similarity.}
\end{enumerate}

The remainder of this paper is organized as follows. In Section \ref{Sec.System Model}, the system model is detailed including transceiver and semantic similarity metrics. 
In Section \ref{Sec.Shared Knowledge Base Construction}, we introduce the shared knowledge base including semantic information theory, the problem formulation and construction algorithm. 
A theoretical analysis of the proposed system, including semantic coding and channel coding, is presented in Section \ref{Sec.Theoretical Analysis} from the view of information theory. Meanwhile, the problem formulation of the proposed semantic communication system is given. 
The realization of the proposed system is detailed in Section \ref{Sec.DNN Implementation}. In Section \ref{Sec.Performance Evaluation}, the performance of our proposed system is shown and compared with benchmarks. Finally, conclusions are summarized in Section \ref{Sec.Conclusions}.

\textit{Notations:} The single boldface letters are used to represent vectors or matrices and single plain capital letters denote integers. Given a vector $\mathbf{x}$, $x_i$ indicates its $i$-th component. The single boldface capital letters denotes random variables and Fraktur capital letters represent sets. $\mathbb{R}^{m \times n}$, $\mathbb{C}^{m \times n}$ represent sets of real and complex matrices of size $m \times n$, respectively. Note that if the superscript is ${(M \times N)\times 1}$, it represents column vectors of size $(M \times N)$ and $1$. $\mathbb{E}(\cdot)$, $\log$ and $\exp$ denotes the expectation, logarithm, exponential function, respectively. $x \sim CN(\mu, \sigma^2)$ means variable follows a circularly-symmetric complex Gaussian distribution with mean $\mu$ and covariance $\sigma^2$.

\begin{figure*}[t]
    \centerline{\includegraphics[scale=0.51]{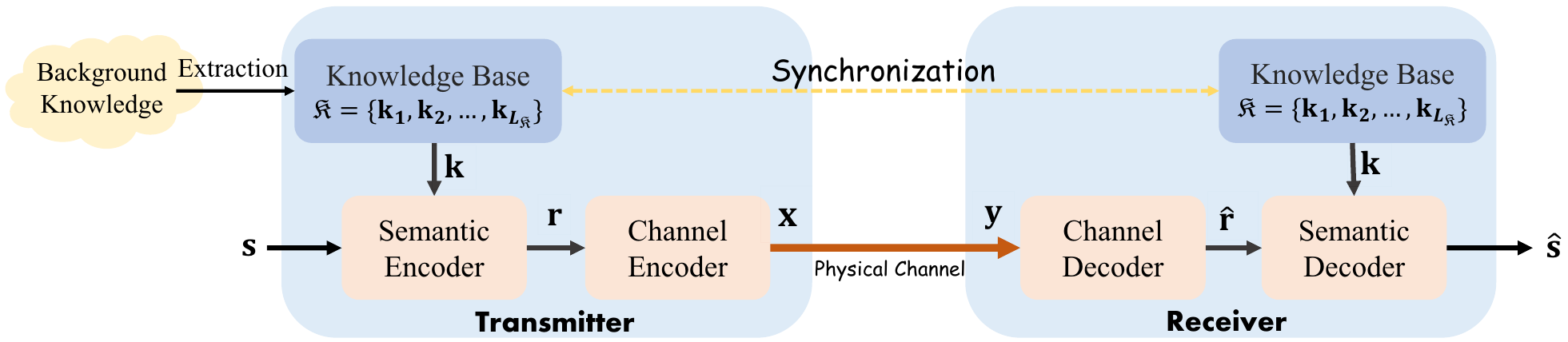}}
    \caption{System model of the proposed semantic communication system.}
    \label{Fig.framework}
    \vspace{-0.2cm}
\end{figure*}

\section{System Model}
\label{Sec.System Model}

Traditional communication coding systems contain two parts: source coding and channel coding. The former is a data compression process with the aim of improving communication effectiveness. The latter is designed to add redundancy appropriately to resist channel noise, which seeks to enhance the reliability of communication. Following this design philosophy, the existing semantic communication systems contain semantic coding and channel coding phases. From another perspective, semantic communication systems comprise three major tasks: 1) semantic information extraction; 2) semantic information compression; 3) physical noise resistance. Unlike previous semantic communication frameworks which utilize end-to-end DNNs to perform the aforementioned tasks at the same time, a novel semantic communication system is proposed for text transmission as shown in Fig. \ref{Fig.framework}, in which a shared knowledge base, constructed by inherently readable sentences, is first introduced in this paper. Through rationally leveraging the knowledge base, the amount of semantic information can be further reduced. 

\subsection{Transmitter}
As shown in Fig. \ref{Fig.framework}, the transmitted message $\mathbf{s}\in \mathbb{R}^{L_s \times 1}$ is a word sequence $[s_1, s_2, s_3, ..., s_{L_s} ]$, which is generated by the source $\mathbf{S}$. Correspondingly, the knowledge base $\mathfrak{K}$, as a set, is constructed by a finite number of knowledge $\mathbf{k}\in \mathbb{R}^{L_k \times 1}$ which is also a word sequence $[k_1, k_2, ..., k_{L_k}]$. 
During the transmission, the message $\mathbf{s}$ is integrated with the knowledge $\mathbf{k}$ at semantic level in semantic encoder $\cal{S(\cdot,\cdot)}$. To minimize the amount of semantic information that needs to be transmitted, $\mathbf{k}$ should be the most similar one to $\mathbf{s}$ at semantic level among the knowledge base $\mathfrak{K}$.
The output of semantic encoder is named as \textbf{residual information} $\mathbf{r}$, which contains semantic information in message $\mathbf{s}$ but not in knowledge $\mathbf{k}$. Intuitively, the more similar $\mathbf{s}$ and $\mathbf{k}$ are, the less information $\mathbf{r}$ contains. 
In the extreme case that $\mathbf{s}$ and $\mathbf{k}$ are the same, $\mathbf{r}$ will not convey any useful information. 

Channel encoder $\cal{C(\cdot)}$ turns vector $\mathbf{r}$ into complex symbols $\mathbf{x}\in \mathbb{C}^{(L_s M) \times 1}$ to cope with the channel attenuation and noise, where $M$ denotes the average number of complex symbols needed for transmitting a single word.
In general, the transmitter maps a message $\mathbf{s}$ into complex symbols $\mathbf{x}$ with the help of knowledge $\mathbf{k}$, which can be written as
\begin{equation}
    \mathbf{x} = \cal{C}(\cal{S}(\mathbf{s},\mathbf{k})).
\end{equation}
Note that according to Slepian-Wolf coding theory \cite{wang2017distributed}, the input knowledge $\mathbf{k}$ at the transmitter can be regarded as side information which is only needed at the receiver. However, in such a case, the semantic encoder must learn the relationship between $\mathbf{S}$ and $\mathbf{K}$ in order to implement effective compression, which poses great challenges in constructing the encoder. On the other hand, the semantic encoder is expected to focus on extracting and fusing semantic information. Therefore, in our system, the knowledge $\mathbf{k}$ is provided to the transmitter and the receiver, which can alleviate the burden of the encoder to learn the relationship between $\mathbf{S}$ and $\mathbf{K}$.

\subsection{Physical Channel}
When the transmitted signal passes through the physical channel, the received signal $\mathbf{y} \in \mathbb{C}^{(L_s M) \times 1}$ can be modeled as
\begin{equation}
    \mathbf{y} = h \mathbf{x} + \mathbf{n}, \label{eq.channel}
\end{equation}
in which $h\in \mathbb{C}$ denotes the channel attenuation and $\mathbf{n}\in \mathbb{C}^{(L_s M)\times 1}$ indicates the noise. 
In this paper, the additive white Gaussian noise (AWGN) channel is adopted, in which $\mathbf{n}$ follows Gaussian distribution and $h$ is equal to $1$. While for fading channels, Rayleigh channel and Rician channel are considered in which $h$ follows $CN(0,1)$ and $CN(\sqrt{\frac{k}{k+1}},\frac{1}{k+1})$, respectively, where $k$ denotes the Rician K factor \cite{DBLP:books/daglib/0091821}.

\subsection{Receiver}
After passing through the physical channel, the received signal $\mathbf{y}$ will be demodulated by the receiver. To be specific, the received signal $\mathbf{y}$ is first processed by the channel decoder $\cal{C}(\cdot)$ to alleviate the effectiveness brought by the physical channel, obtaining the estimated residual information $\hat{\mathbf{r}}$. Similar to the semantic encoder, the semantic decoder has two inputs, i.e., the estimated residual information $\hat{\mathbf{r}}$ and the corresponding knowledge $\mathbf{k}$. For convenience, we denote the channel decoder and semantic decoder as $\mathcal{C}^{-1}(\cdot)$ and $\mathcal{S}^{-1}(\cdot,\cdot)$, respectively.
The decoded process can be represented as 
\begin{equation}
    \hat{\mathbf{s}}= \mathcal{S}^{-1}(\mathcal{C}^{-1}(\mathbf{y}),\mathbf{k}),
\end{equation}
where $\hat{\mathbf{s}}$ is the recovered sentence. Note $\cal{S(\cdot,\cdot)}$, $\cal{C(\cdot)}$, $\mathcal{C}^{-1}(\cdot)$ and $\mathcal{S}^{-1}(\cdot,\cdot)$ have different weights. 

In our semantic communication framework, the knowledge base at the receiver side is the same as that at the transmitter side, which indicates that they always contain the same background knowledge. This can be easily achieved by generating a knowledge base in a could server and broadcasting a copy of the knowledge base to each user. Since the knowledge base is constructed by a limited number of sentences referred to as knowledge, each of them is assigned a unique index. During the transmission, the index of the used knowledge is passed to the receiver through the control plane \cite{3GPP} without errors, enabling the receiver to refer to the same knowledge.

\subsection{Semantic Similarity Metrics}
In traditional communication systems, bit error rate (BER) and symbol error rate (SER) are commonly used as performance metrics. For semantic communications, however, neither BER nor SER cannot accurately evaluate the difference between two sentences at the semantic level, since a few bit errors may also incur severe misunderstand of the whole sentence. In order to tackle this issue, bilingual evaluation understudy (BLEU) \cite{DBLP:conf/acl/PapineniRWZ02} is adopted in this paper. BLEU is widely used for evaluating machine translation and has been applied to recently emerged semantic communication systems \cite{DBLP:journals/tsp/XieQLJ21}. BLEU computes the $n$-gram similarity of two given sentences, based on the linguistic law that semantically consistent words usually come together in a given corpus. The $n$-gram denotes a word group set with each element is $n$ words, which is constructed by splitting the given sentence with overlap. The BLEU score is calculated by 
\begin{equation}
    BLEU = \min(1, \exp(1- \frac{len(\hat{\mathbf{s}})}{len(\mathbf{s})})) \times \exp{(\sum_{n=1}^N \omega_n \log{p_n})},
    \label{Eq.BLEU}
\end{equation}
in which $len(\cdot)$ denotes the length of a given sentence. $\omega_n$ denotes the weight of $n$-gram and $p_n$ is the $n$-gram score, which is formulated as
\begin{equation}
    p_n = \frac{\sum_{l_n \in \hat{\mathbf{s}}} \min (C_{l_n}(\hat{\mathbf{s}}), C_{l_n} (\mathbf{s})) }{\sum_{l_n \in \hat{\mathbf{s}}} C_{l_n}(\hat{\mathbf{s}})},
\end{equation}
where $l_n$ is one of elements in the $n$-gram and $C_{l_n}(\hat{\mathbf{s}})$ is the frequency count function for the occurrence of $l_n$ in $\hat{\mathbf{s}}$.

However, since BLEU cannot recognize synonym and polysemy, a DL-powered metric, named as sentence similarity score \cite{DBLP:journals/tsp/XieQLJ21}, is adopted in our system, which is formulated as 
\begin{equation}
    SS (s,\hat{s}) =  \frac{\mathcal{B}_{\phi}(s)\cdot\mathcal{B}_{\phi}{(\hat{s})}^T}{||\mathcal{B}_{\phi}(s)||~||\mathcal{B}_{\phi}(\hat{s})||},
    \label{Eq.SS}
\end{equation}
in which $\mathcal{B}_{\phi}(\cdot)$ represents the sentence transformer model \cite{DBLP:conf/emnlp/ReimersG19} converting a sentence into a high dimensional vector. Since $\mathcal{B}_{\phi}(\cdot)$ is pre-trained by millions of data on semantic textual matching tasks, it is reasonable to assume that the semantic information of sentences is separable in the high dimensional space. The degree of similarity between two vectors is then calculated by cosine similarity and named as $Similarity~Score$, which yields a value ranging from zero to one.

\begin{figure*}[t]
    \centerline{\includegraphics[scale=0.62]{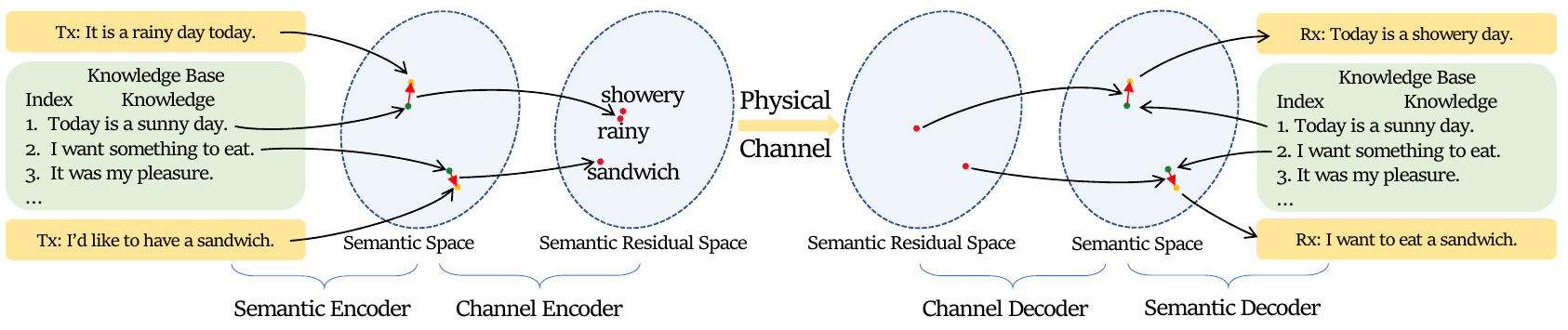}}
    \caption{The theoretical analysis of our proposed semantic communication system with a shared knowledge base.}
    \label{Fig.theory}
\end{figure*}

\section{Shared Knowledge Base Construction}
\label{Sec.Shared Knowledge Base Construction}

In this section, we first elaborate the shortcomings of the existing semantic information theory. To alleviate these issues, the shared knowledge base is first introduced into the semantic communication system. Moreover, semantic self-information and entropy are mathematically defined based on the knowledge base, which guides the design of semantic communication system. As an essential component of the system, the problem formulation of the knowledge base construction is given. Finally, the shared knowledge base construction algorithm is detailed by viewing the problem as a clustering task. 

\subsection{Semantic Information Theory}
In Shannon's information theory, the message is considered a sequence of letters, and the entropy of the source is characterized by the probability of each letter occurring. The implicit assumption is that letters are orthogonal to each other, which is true since the semantic aspect is ignored. 
Following the philosophy of Shannon's information theory, in \cite{DBLP:journals/cn/StrinatiB21}, the semantic entropy of the source is defined by the occurrence probability of a valid meaning, which implies different sentences may have the same meaning since the presence of synonyms and sentence structure. However, different from Shannon's information theory, the meanings of sentences are not orthogonal. To be specific, different sentences may have part of meanings in common. For example, the sentence ``Today is a sunny day'' and another sentence ``It is a rainy day today'' are slightly different sentences and do not express exactly the same meaning, but it is clear that they mean partly the same thing, both being descriptions of the weather today. Therefore, the direct use of occurrence probabilities of different semantic meanings to represent the semantic self-information may be inappropriate, which ignores the semantic relevance between different sentences. To alleviate this issue, a world model is introduced in \cite{bao2011towards}, which decomposes semantic meanings of sentences into combinations of interpretations. Nevertheless, it is a very tricky thing to implement the complex world model and the precise logical reasoning mechanisms.

In our semantic communication system, with the aid of the knowledge base that can be seen as a simple world model, the semantic self-information of the message $I_s (\mathbf{s})$ and semantic entropy of the source $H_s(\mathbf{S})$ can be derived. 
Specifically, the meaning of each message can be divided into two parts, which are the information in common with the knowledge base, i.e., the corresponding knowledge $\mathbf{k^*_s}$, and the information different from the identified knowledge, i.e., the residual information $\mathbf{r}$. 
It is noteworthy that there is no overlap of semantic information between $\mathbf{k^*_s}$ and $\mathbf{r}$. Meanwhile, given an arbitrary knowledge $\mathbf{k}$ in $\mathbf{K}$, the residual information cannot be identified without message $\mathbf{s}$, and vice versa. 
Hence, $\mathbf{k^*_s}$ and $\mathbf{r}$ can be regarded as mutually independent in semantic sense, which means $I_s (\mathbf{s})$ is equal to the sum of the semantic self-information of $\mathbf{k^*_s}$ and $\mathbf{r}$.

To formulate the semantic self-information of $\mathbf{k}$, we assume that each piece of knowledge in the knowledge base represents a commonly used meaning, and any two pieces of knowledge are irrelevant, which is guaranteed by Algorithm \ref{alg.kb}. Moreover, the size of the knowledge base is finite. In such a case, the self-information of each piece of knowledge and the entropy of the knowledge base can be formulated in Shannon's sense, which are given as 
\begin{equation}
    I(\mathbf{k}) = -\log p(\mathbf{k}),
\end{equation}
\begin{equation}
    H(\mathbf{K}) = - \sum_{\mathbf{k}\in \mathfrak{K}} p(\mathbf{k}) \log (p(\mathbf{k})),
\end{equation}
respectively, where $p(\mathbf{k})$ is the probability of knowledge $\mathbf{k}$ among the knowledge base, and $\mathbf{k}$ denotes a random piece of knowledge, as a random variable. 

Furthermore, since the residual information $\mathbf{r}$ is obtained by the semantic encoder with the aid of $\mathbf{k^*_s}$, the knowledge that is most similar to the message $\mathbf{s}$ at semantic level must be first identified by 
\begin{equation}
    \mathbf{k}^{*}_{\mathbf{s}} = \mathop{\arg\min}\limits_{\mathbf{k}\in \mathfrak{K}} D(\mathbf{s},\mathbf{k}),
    \label{Eq.k}
\end{equation}
where $D(\mathbf{s},\mathbf{k})$ represents the semantic distance between the message $\mathbf{s}$ and the knowledge $\mathbf{k}$, ranging from zero to one. 
Different from the knowledge $\mathbf{k}$ in $\mathbf{K}$, the residual information is not orthogonal to each other (there exists a semantic overlap) and the set of residual information is infinite. Hence, the self-information of $\mathbf{r}$ cannot be defined by probability theory. 
To address this issue, the degree of difference between $\mathbf{s}$ and $\mathbf{k}^{*}_{\mathbf{s}}$ at the semantic level is utilized to evaluate the amount of residual information. The more dissimilar $\mathbf{s}$ and $\mathbf{k}^{*}_{\mathbf{s}}$ are, the more the amount of residual information is. 
Therefore, the semantic self-information of the message $I_s (\mathbf{s})$ consists of the most similar knowledge $\mathbf{k}^{*}_{\mathbf{s}}$ and the distance between the message and the most similar knowledge, and it can be defined as
\begin{equation}
    I_s(\mathbf{s}) = I(\mathbf{k}^{*}_{\mathbf{s}}) - \lambda \log [1-D(\mathbf{s},\mathbf{k}^{*}_{\mathbf{s}})],  
    \label{Eq.I_s(s)}
\end{equation}
where $\lambda$ is a positive number to balance the quantization relation between the first term and the second term. According to \eqref{Eq.k}, the term $(1 - D(\mathbf{s},\mathbf{k}^{*}_{\mathbf{s}}))$ stands for the sentence similarity, which can be evaluated by using BLEU \eqref{Eq.BLEU} or sentence similarity score \eqref{Eq.SS}. 
When $\mathbf{s}=\mathbf{k_s}$, the semantic distance between the message $\mathbf{s}$ and the knowledge $\mathbf{k_s}$ approaches zero, resulting in $D(\mathbf{s},\mathbf{k_s}) = 0$ and consequently $I_s(\mathbf{s}) = I(\mathbf{k_s})$. 
Moreover, given a corpus $\mathfrak{S}$, the semantic entropy of the source can be driven from \eqref{Eq.I_s(s)} and defined as
\begin{equation}
    H_s(\mathbf{S}) = \sum_{\mathbf{s}\in \mathfrak{S}} p(\mathbf{s}) I_s (\mathbf{s}),
\end{equation}
where $p(\mathbf{s})$ denotes the occurrence probability of the transmitted message $\mathbf{s}$ and satisfies $\sum_{\mathbf{s}\in \mathfrak{S}} p(\mathbf{s}) = 1$. 
It is worth noting that the calculation of semantic entropy depends on both prior knowledge and the given corpus due to the fact that semantics cannot exist independently from the prior information. Similarly, in \cite{bao2011towards}, the semantic entropy is determined by both the corpus and a world model, which contains background knowledge. 
In such cases, the semantic self-information and semantic entropy can be well defined and are of guidance for the construction of knowledge base and the implementation of semantic communication systems.

\subsection{Problem Formulation}
Overall, the knowledge base, as a simple world model, is of great importance and is expected to be constructed as the preliminary for the proposed semantic communication system. According to \eqref{Eq.I_s(s)}, the semantic meanings of message $\mathbf{s}$ can be divided into $I(\mathbf{k}^{*}_{\mathbf{s}})$ and $- \lambda \log [1-D(\mathbf{s},\mathbf{k}^{*}_{\mathbf{s}})]$. Due to the limitation of knowledge base size, for any transmitted message $\mathbf{s}$, a good knowledge base $\mathfrak{K}$ tends to maximize the corresponding $I(\mathbf{k}^{*}_{\mathbf{s}})$. Moreover, the knowledge base must satisfy two constraints, which means each piece of knowledge stands for a commonly used meaning in the corpus and is irrelevant to each other. 
On the one hand, the collected dataset (background knowledge) is leveraged for constructing the knowledge base, i.e., $\mathfrak{K}=\{\mathbf{k_1}, \mathbf{k_2}, ..., \mathbf{k}_{lk}\}$, ensuring that each piece of knowledge represents a valid meaning. Through clustering, the one that is commonly used can be identified.
On the other hand, $similarity~score$ in \eqref{Eq.SS} and a pre-configured threshold $\theta$ are adopted to determine orthogonality between sentences. Generally, two sentences are considered to be irrelevant if their $similarity~score$ is less than the threshold $\theta$, and vice versa. Hence, the optimization problem for the construction of knowledge base can be formulated in the following.

\begin{myOpt}
    \setlength{\abovedisplayskip}{10pt}
    \setlength{\belowdisplayskip}{1pt}
    In this paper, the objective of knowledge base construction is to minimize the expectation of the distance between the message $\mathbf{s}$ and its most similar knowledge $\mathbf{k}^{*}_\mathbf{s}$ identified by \eqref{Eq.k} among the knowledge base $\mathfrak{K}$, i.e.,
    \begin{align}
        \label{Eq.P1}
        &\mathop{\min}\limits_{\mathfrak{K}} \mathop{\mathbb{E}}\limits_{\mathfrak{s}} [D(\mathbf{s},\mathbf{k}^{*}_\mathbf{s})],\\
        &
        \begin{aligned}
            ~\mathrm{s.t.} & \quad lk \leq L_K, \\
            & \quad D(\mathbf{k}_i,\mathbf{k}_j)\leq \theta, \forall i,j \in [1, ..., lk]~\textrm{and}~i \neq j
        \end{aligned}
    \end{align}
    where $L_K$ represents the maximum size of the knowledge base.
\end{myOpt}

\subsection{Shared Knowledge Base Construction Algorithm}
As mentioned above, the shared knowledge base is a critical part in our semantic communication system. In this subsection, the construction of the shared knowledge base is detailed. Generally, the knowledge base is composed of elements in real world, stored as sentences or triplets \cite{DBLP:journals/cm/ShiXLX21}. In our system, sentences with more semantic information are adopted as basic atoms instead of triplets to build the knowledge base.

As stated above, \emph{\textbf{Problem} 1} can be viewed as a clustering task, which establishes a knowledge base. Nevertheless, the commonly used clustering methods, such as k-means, are inappropriate for this problem. 
On the one hand, the semantic representations of messages are high-dimensional vectors. On the other hand, the collected dataset (background knowledge) that needs to be clustered is vast, and the number of cluster centers is large. Performing clustering tasks on a huge amount of high-dimensional data is a tricky thing.
To address thees issues, an algorithm for the construction of knowledge base is developed based on the single pass clustering algorithm \cite{jo2008single} as shown in Algorithm \ref{alg.kb}.

Specifically, given a sentence in the dataset, it will be compared in sequence with the knowledge in the current knowledge base to obtain a $similarity~score$.
If the computed scores are larger than $\theta$, the sentence and the knowledge are believed to have similar semantic meanings, which means there is no need to add the sentence to the knowledge base. In contrast, if all $similarity~scores$ between the sentence and all knowledge are smaller than $\theta$, the sentence is considered to represent a new semantic meaning and will be incorporated into the knowledge base.
Therefore, the threshold $\theta$ has the ability to control the size of knowledge base.
As $\theta$ grows to one, more sentences will be absorbed into the knowledge base, resulting in a richer knowledge base, which requires longer running time and larger storage size.
Conversely, when $\theta$ is going down to zero, the size of the knowledge base decreases while the amount of residual information that needs to be transmitted increases.
In our experiment, $\theta$ is manually set to balance the size of knowledge base and the transmission data size.

It should be noted that the transmitter and receiver share the same knowledge base and each sentence is assigned a unique index. Since the size of knowledge base is acceptable (e.g., about one thousand sentences are selected as the knowledge out of over $270k$ training data when $\theta$ is $0.3$ in our simulation), the index requires only a few bits to represent. Thus, it is reasonable to assume that the index of the adopted knowledge is sent to the receiver through control plane \cite{3GPP} without errors. 
Moreover, it should be used to update the knowledge base by incorporating newly collected data. More specifically, a cloud server could identify new knowledge based on the existing knowledge base and broadcast the newly added portions to all users. At the cloud server, Algorithm \ref{alg.kb} can be utilized to update the knowledge base with minor modifications. In this case, the initialized knowledge base would be the existing knowledge base rather than an empty set. 

\begin{algorithm}[t]
    \caption{Construction of Knowledge Base.}
    \begin{algorithmic}[1]  
        \label{alg.kb}
        \REQUIRE {Training data set $\mathfrak{S}$.}
        \ENSURE {Knowledge base $\mathfrak{K}$ and a threshold $\theta$}
        \STATE Initialize the knowledge base $\mathfrak{K}$ to an empty set.
        \FOR {$\mathbf{s}_i$ in $\mathfrak{S}$}
            \IF {$\mathfrak{K}$ is empty}
            \RETURN Step 12 
            \ENDIF
            \FOR{$\mathbf{k}_j$ in $\mathfrak{K}$}
            \STATE {$Similarity~Score \leftarrow \mathcal{SS} (\mathbf{s}_i,\mathbf{k}_j)$ according to (6)}
            \IF {$Similarity~Score > \theta$}
            \RETURN Step 2 
            \ENDIF
        \ENDFOR
        \STATE {Add $\mathbf{s}_i$ into  $\mathfrak{K}$}
        \ENDFOR
    \end{algorithmic}
\end{algorithm}

\section{Theoretical Analysis of Semantic Communication System with a Shared Knowledge Base}
\label{Sec.Theoretical Analysis}

In this section, motivated by the semantic information theory, a theoretical analysis of the proposed system is presented.
To be specific, we elaborate on semantic encoder and channel encoder in semantic communication system, respectively, and how they bring a considerable performance gain.
Finally, the problem formulation of the proposed system is given based on the shared knowledge base.

\subsection{Semantic Encoder and Decoder}

Motivated by the semantic information theory, an example of transmission in semantic communication systems can be drawn in Fig. \ref{Fig.theory}.
At the semantic encoder, the transmitted message as well as the corresponding knowledge are transformed by the semantic encoder into points in a high-dimensional semantic space. Sentences with similar meanings should have a small distance in the semantic space and vice versa. 
Using basic information-theoretic tools, the semantic entropy of the source, $H_s(\mathbf{S})$, can be written as
\begin{equation}   
    H_s (\mathbf{S}) = H(\mathbf{K}) + H_s (\mathbf{S}/\mathbf{K}) - H_s (\mathbf{K}/\mathbf{S}),
    \label{Eq.H_s(S)}
\end{equation}
where $H_s (\mathbf{K}/\mathbf{S})$ denotes the semantic entropy of $\mathbf{K}$ conditioned to $\mathbf{S}$ and $H_s (\mathbf{S}/\mathbf{K})$ denotes the semantic entropy of $\mathbf{S}$ conditioned to $\mathbf{K}$. 
According to \eqref{Eq.k}, for each message, there is a unique piece of knowledge in the knowledge base corresponding to it, which means $H_s (\mathbf{K}/\mathbf{S}) = 0$. Besides, we use \textit{residual information} $\mathbf{r}$ to represent the obtained information that contained in $\mathbf{s}$ but not in $\mathbf{k}$. Normally, the semantic entropy of residual information, $H_s (\mathbf{R})$ is not greater than $H_s (\mathbf{S}|\mathbf{K})$, due to the fact that only a finite number of complex symbols are used to transmit information. Under ideal circumstances, $H(\mathbf{R})$ is equal to $H_s (\mathbf{S}/\mathbf{K})$. Then, \eqref{Eq.H_s(S)} can be further written as
\begin{equation}
    H_s (\mathbf{S}) = H(\mathbf{K}) + H_s (\mathbf{R}),
\end{equation}
which means the emitted $\mathbf{s}$ can be accurately estimated by the current knowledge $\mathbf{k}$ and the residual information $\mathbf{r}$ that is contained in $\mathbf{s}$ but not in $\mathbf{k}$. In such a case, with the aid of the aforementioned knowledge base, the amount of source information that needs to be transmitted, i.e., residual information, is greatly reduced since $H_s (\mathbf{R}) < H_s (\mathbf{S})$, which justifies the compression space of our system. Therefore, the aim of semantic encoder is to obtain residual information, $\mathbf{r}$, indicated by the red arrow in the semantic space as shown in Fig. \ref{Fig.theory}.

At the receiver side, the semantic decoder converts the knowledge $\mathbf{k}$ into a vector in the semantic space. The received residual information $\hat{\mathbf{r}}$ is then combined with this vector to obtain the estimated message $\hat{\mathbf{s}}$. If $\hat{\mathbf{r}}$ is polluted by the noise, the received message $\hat{\mathbf{s}}$ may have some errors. Nevertheless, even though the received residual information $\hat{\mathbf{r}}$ is correct, the syntax of the received message $\hat{\mathbf{s}}$ may change due to the limited transmission symbols. In such a case, the semantic meaning of the transmitted message may remain unchanged, as shown in Fig. \ref{Fig.theory}.

\subsection{Channel Encoder and Decoder}

In conventional communication systems, a sequence of bits encoded by a source is considered semantically meaningless. Conventional channel encoders thus only consider the error-free transmission over noisy channels, resulting the transmission rate limited by the channel capacity. While in semantic communication systems, the input of channel encoder is residual information with practical meanings. 
Channel coding requires not only adding redundancy to the residual information enhancing noise resistance ability, but also maintaining a semantic distance in symbol space which is similar to that in the semantic space. To be specific, the channel encoder maps the residual information $\mathbf{r}$ into complex symbols $\mathbf{x}$, in which residual information with similar meanings is mapped to symbols with smaller inter-symbol distances. In such a case, the meaning of messages may be unaffected when some errors occur. For example, symbols with meanings of ``rainy'' and ``showery" are close to each other in semantic residual space space as shown in Fig. \ref{Fig.theory}. During the transmission of symbols with meaning of ``rainy'', if an error occurs due to the physical noise and can not be eliminated by the channel decoder, the received residual information could be ``showery'', keeping the meaning of the transmitted message unchanged. Generally, by adopting channel coding, the semantic meanings of the sent message can be delivered accurately even when the errors cannot be eliminated entirely.

\subsection{Problem Formulation}

In this paper, a DNN-enabled semantic encoder performs the fusion of $\mathbf{s}$ and $\mathbf{k}$, obtaining the residual information $\mathbf{r} \in \mathbf{R}$. Then the channel encoder based on DNNs maps residual information $\mathbf{r}$ into complex symbols $\mathbf{x}$. After passing through the physical channel, the received complex symbols $\mathbf{y}$ are converted by the channel decoder into the estimated residual information $\hat{\mathbf{r}}$. Finally, the received message $\hat{\mathbf{s}}$ can be obtained by combining $\hat{\mathbf{r}}$ and $\mathbf{k}$. 
To train DNNs in an end-to-end manner given the construction of knowledge base, the objective of our semantic communication system is to emit the least complex symbols $\mathbf{x}$, i.e., $H_s (\mathbf{R}) \leq H_s (\mathbf{S}|\mathbf{K})$, and reconstruct the source message accurately at the destination side. Specifically, for a given message $\mathbf{s}$ from source $\mathbf{S}$, the probability $p(\hat{\mathbf{s}}=\mathbf{s})$ should be maximized.
Equivalently, the maximization problem can be converted to a minimization problem formulated in the following.
\begin{myOpt}
    \setlength{\abovedisplayskip}{10pt}
    \setlength{\belowdisplayskip}{1pt}
    In this paper, the objective of the semantic communication system is to accurately recover the semantic meaning of the transmitted message at the receiver side. Mathematically, the objective can be formulated by minimizing the expectation of the distance between the transmitted message $\mathbf{s}$ and the reconstructed message $\hat{\mathbf{s}}$, i.e., 
    \begin{align}
        \label{eq.minD}
        &\min_{\mathcal{{S}(\cdot,\cdot)}, \mathcal{S}^{-1}(\cdot,\cdot), \mathcal{{C}(\cdot)},  \mathcal{C}^{-1}(\cdot)} \sum_{\mathbf{s}} \sum_{\hat{\mathbf{s}}} p(\mathbf{s}, \hat{\mathbf{s}}) d(\mathbf{s}, \hat{\mathbf{s}}), \\
            & \begin{array}{r@{\quad}r@{}l@{\quad}l}
              \quad \quad \quad ~~ \mathrm{s.t.}  & \quad \quad H_s (\mathbf{R}) \leq H_s (\mathbf{S}|\mathbf{K}),\\
          \end{array}
      \end{align}
    where $d(\mathbf{s}, \hat{\mathbf{s}})$ represents the distance measurement function of sentences $\mathbf{s}$ and $\hat{\mathbf{s}}$. 
\end{myOpt}

Generally, $d(\cdot, \cdot)$ can be achieved by various measurements according to the granularity. Specifically, conventional communication systems consider bit error rate (BER) and symbol error rate (SER) as loss functions. Nevertheless, the granularity of these measurements is too small for semantic communications. Moreover, $similarity~score$ \eqref{Eq.SS} is not suitable for the practical implementation as a loss function due to its inherent complexity, which may lead to issues such as gradient vanishing and gradient explosion. 
Additionally, BLEU \eqref{Eq.BLEU} involves counting operations which are not derivable. Therefore, in order to achieve successful recovery at the semantic level, differences between words are more appropriate as our goal than differences in bits.
To be specific, the cross entropy (CE) loss, $\mathcal{L}_{CE}$, is leveraged as the distance function $d(\mathbf{s},\hat{\mathbf{s}})$ to calculate differences between the original message $\mathbf{s}$ and the estimated one, denoted as
\begin{equation}
    \begin{aligned}
    \mathcal{L}_{CE}  = - \sum_{\omega \in \mathbf{s}} q(\omega) \log (p(\omega)) + (1-q(\omega)) \log ( 1 - p (\omega) ),
    \label{eq.ce}
    \end{aligned}
\end{equation}
where $q(\omega)$ and $p(\omega)$ denote the probability of the word $\omega$ that appears in sentence $\mathbf{s}$ and $\hat{\mathbf{s}}$, respectively.
Since the differences between two probability distributions can be measured by the CE loss, the DNN enabled encoder is optimized to learn the word distribution by minimizing the expectation of the distance defined in \eqref{eq.minD}.

\begin{figure*}[t]
    \centerline{\includegraphics[scale=0.6]{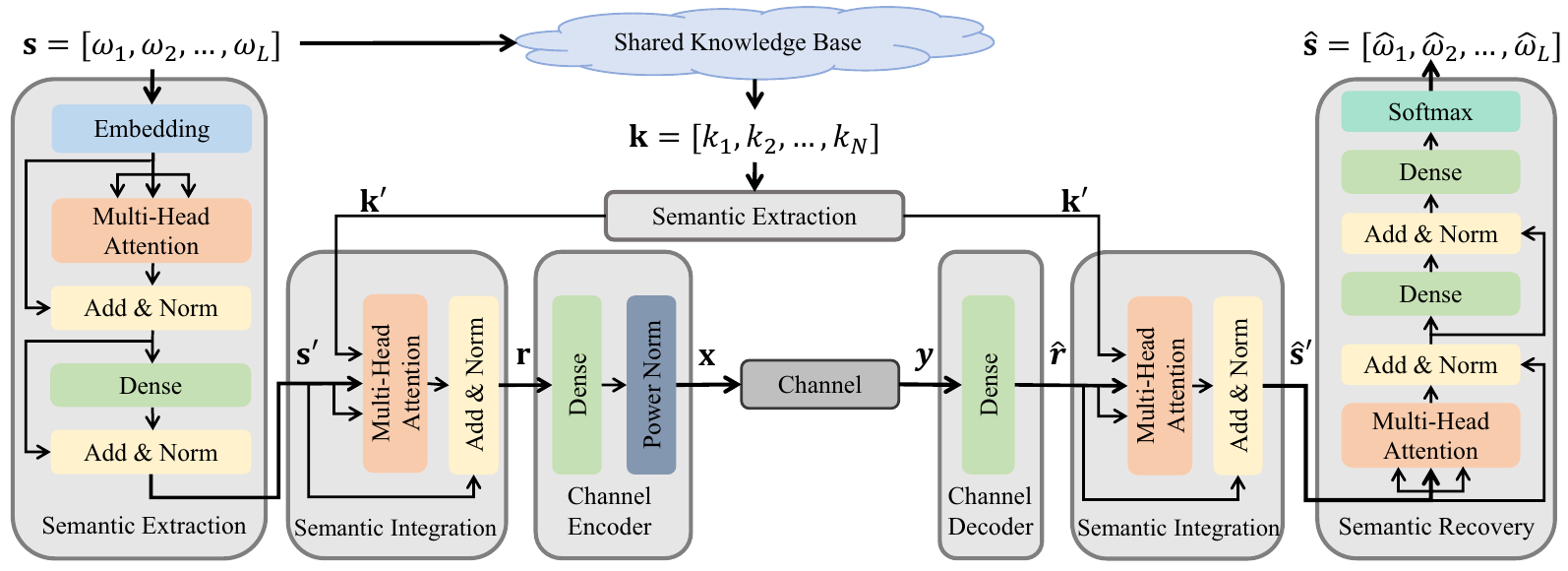}}
    \caption{The network structure of our proposed semantic communication system driven by knowledge base.}
    \label{Fig.structure}
    \vspace{-1em}
\end{figure*}

\section{DNN Implementation of Semantic Communication System with a Shared Knowledge Base}
\label{Sec.DNN Implementation}

In this section, the network structure of the proposed system is detailed and shown in Fig. \ref{Fig.structure}. First, the self-attention mechanism and multi-head attention layer are described, as they are crucial to our network. Next, the structure of the semantic encoder is detailed including semantic extraction module and semantic integration module. A discussion of channel encoder and decoder is followed. Semantic decoder is then presented as the final part of the network.
Additionally, the training process is illustrated in Algorithm \ref{alg.train} where the transmitter and receiver are jointly optimized to achieve successful transmission.

\subsection{Self-attention Mechanism}
Before detailing the structure of our network, we first introduce self-attention mechanism and multi-head attention layer \cite{DBLP:conf/nips/VaswaniSPUJGKP17} as preliminary, which are often used to process sequentially data.
In natural language processing, a word in a sentence is not independent and has dependencies with the context. Self-attention mechanism works by capturing correlations between vectors, which enables the network to see the global information and focus on key information.

Specifically, the mathematical expression of self-attention mechanism, i.e., the attention function, can be formulated by
\begin{equation}
    Attention (\mathbf{Q_{A},K_{A},V_{A}}) = Softmax (\frac{\mathbf{Q_{A}} \mathbf{W}_Q {( \mathbf{K_{A}} \mathbf{W}_K)}^{T}}{\sqrt{d}}) \mathbf{V_{A}} \mathbf{W}_V,
    \label{eq:attention}
\end{equation}
where $\mathbf{W}_Q \in \mathbb{R}^{E \times d}$, $\mathbf{W}_K \in \mathbb{R}^{E \times d}$ and $\mathbf{W}_V \in \mathbb{R}^{E \times d}$ are the three learnable weight matrices, and $d$ is the dimension used in the attention function.
The inputs of the function is query, $\mathbf{Q_{A}}\in \mathbb{R}^{L \times E}$, key, $\mathbf{K_{A}}\in \mathbb{R}^{L \times E}$ and value, $\mathbf{V_{A}} \in \mathbb{R}^{L \times E}$.
In some cases, the input data for $\mathbf{Q_{A}}$, $\mathbf{K_{A}}$, and $\mathbf{V_{A}}$ are the same.
The $Softmax(\cdot)$ function \cite{DBLP:books/aw/RN2020} is used to convert the input vector into a normalized probability distribution.
In such a case, the normalized intensity of attention that a given vector pays to any vector can be obtained, i.e., $Softmax (\frac{\mathbf{Q_{A}} \mathbf{W}_Q {( \mathbf{K_{A}} \mathbf{W}_K)}^{T}}{\sqrt{d}})$.
Finally, the output is the matrix product of the attention matrix, $Softmax (\frac{\mathbf{Q_{A}} \mathbf{W}_Q {( \mathbf{K_{A}} \mathbf{W}_K)}^{T}}{\sqrt{d}})$, and $\mathbf{V_{A}} \mathbf{W}_V$ representing the value of the each vector.

Intuitively, the $Attention(\cdot,\cdot,\cdot)$ function projects data into a certain subspace and finds its correlations. To enhance the ability of capturing features, the multi-head attention function, as an extention of $Attention(\cdot,\cdot,\cdot)$ function, projects data into multiple subspaces to achieve better performance.
Specifically, $MultiHead(\cdot,\cdot,\cdot)$ function is given as
\begin{align}
    MultiHead (\mathbf{Q_{A},K_{A},V_{A}}) = Concat (\mathbf{Z}_1,...,\mathbf{Z}_h) \mathbf{W}_o, \\
    \mathrm{where}~\mathbf{Z}_i = Attention(\mathbf{Q_{A},K_{A},V_{A}}),
    \label{eq:multihead}
\end{align}
where $\mathbf{W}_o \in \mathbb{R}^{(d h) \times E}$ denotes a learnable weight matrix, $h$ is the number of subspaces, and $Concat(\cdot)$ concatenates the given sequence of matrices in a specific dimension. It should be noted that the output matrix shares the same shape as the input matrix.

\subsection{Semantic Encoder}
According to different functions, the semantic encoder is spilt into two modules, i.e., semantic extraction module, which performs feature extraction facilitating subsequent processing, and semantic integration module, which fuses the processed message and knowledge. 

\subsubsection{Semantic Extraction Module}
As shown in Fig. \ref{Fig.structure}, the input of network is the transmitted sentence $\mathbf{s} \in \mathbb{R}^{L_s\times 1}$, where each element is an integer representing a specific word in the codebook and $L_s$ is the length of the sentence. 
In order to keep the words relatively independent while reducing computational complexity, the input sentence is embedded first obtaining $\mathbf{s}_1 \in \mathbb{R}^{L_s\times E}$ where $E$ is the embedding dimension. 
The embedded matrix $\mathbf{s}_1$ is then fed into a multi-head attention layer as both a query, a key and a value, which is give as 
\begin{equation}
    \mathbf{s}_2 = MultiHead (\mathbf{s}_1,\mathbf{s}_1,\mathbf{s}_1).
\end{equation}
To avoid the problem of vanishing and exploding gradient, a residual connection is leveraged across the multi-head attention layer followed by a normalization layer, which is given by
\begin{equation}
    \mathbf{s}_3  = Norm (\mathbf{s}_1 + \mathbf{s}_2),
\end{equation}
where $Norm(\cdot)$ represents layer normalization method \cite{DBLP:journals/corr/BaKH16}, normalizing each feature of activations to a zero mean and unit variance.

After the above processing, a dense layer, including two fully connected layers and an activation function $GELU(\cdot)$ \cite{DBLP:journals/corr/HendrycksG16}, is employed to enhance the non-linear processing capability of the network. Specifically, the data is projected into a higher dimensional space and then projected back into the original dimensional sapce after a non-linear process. Besides, a residual connection and a normalization layer are leveraged. In general, the process is given by
\begin{equation}
    \mathbf{s'} = Norm (\mathbf{s}_3 + (GELU(\mathbf{s}_3 \mathbf{W}_1 + \mathbf{b}_1)\mathbf{W}_2 + \mathbf{b}_2 ),
\end{equation}
where $\mathbf{W}_1\in \mathbb{R}^{E\times 4E}$ and $\mathbf{W}_1\in \mathbb{R}^{4E\times E}$ are two weight matrices, and $\mathbf{b}_1 \in \mathbb{R}^{L_s\times 4E}$ and $\mathbf{b}_1 \in \mathbb{R}^{L_s\times E}$ are bias.
Finally, the output of semantic extraction module is $\mathbf{s'} \in \mathbb{R}^{L_s\times E}$, which represents the semantic information of the input message $\mathbf{s} \in \mathbb{R}^{L_s\times 1}$.

In particular, the knowledge stored in the form of sentences also needs to be converted into vectors that characterize semantic information. Therefore, in order to reduce the size of the model and the complexity of the network, the semantic extraction module is also used to deal with knowledge, obtaining $\mathbf{k'}\in \mathbb{R}^{L_k\times E}$, which shares the same parameters with that for processing messages.

\subsubsection{Semantic Integration Module}
After obtaining $\mathbf{s'}$ and $\mathbf{k'}$, the next objective is to acquire residual information, $\mathbf{r}$, that contains information in message but not in knowledge. 
To be specific, self-attention mechanism is adopted in semantic integration module, where the inputs of the multi-head attention layer are $\mathbf{k'}$ as query, and $\mathbf{s'}$ as key and value, given by 
\begin{equation}
    \mathbf{{s}'}_1 = MultiHead (\mathbf{k'},\mathbf{s'},\mathbf{s'}).
\end{equation}
According to \eqref{eq:attention}, the matrix product of $\mathbf{k'} \mathbf{W}_Q$ and $\mathbf{s'} \mathbf{W}_K$ represents the semantic relationship between knowledge and message. After the normalization process, $Softmax (\frac{\mathbf{k'} \mathbf{W}_Q {( \mathbf{s'} \mathbf{W}_K)}^{T}}{\sqrt{d}})$ denotes the degree to which the semantic component in $\mathbf{s'}$ is present in $\mathbf{k'}$. Then, by multiplying by $\mathbf{s'} \mathbf{W}_V$, the semantic component of $\mathbf{k'}$ contained in $\mathbf{s'}$ can be reduced. Finally, a residual connection and a normalization layer are followed to further process $\mathbf{{s}'}_1$ and acquire residual information $\mathbf{r'}\in \mathbb{R}^{L_s\times E}$.

\begin{algorithm}[t]
    \caption{Training of the Proposed Neural Network.}
    \begin{algorithmic}[1] 
        \REQUIRE {Training data set $\mathfrak{S}$ and knowledge base $\mathfrak{K}$}
        \ENSURE { The whole network $\mathcal{{S}(\cdot,\cdot)}$, $\mathcal{S}^{-1}(\cdot,\cdot)$, $\mathcal{{C}(\cdot)}$ and $\mathcal{C}^{-1}(\cdot)$}
        \FOR {$\mathbf{s}_i$ in $\mathfrak{S}$}
            \STATE {$\mathbf{k}_j$ $\leftarrow$ Find the knowledge with respect to the most similar score to $\mathbf{s}_i$ in $\mathfrak{K}$}
            \STATE {$\mathbf{r}_i \leftarrow \mathcal{S}(\mathbf{s}_i, \mathbf{k}_j)$}
            \STATE {$\mathbf{x}_i \leftarrow \mathcal{C}(\mathbf{r}_i) $}
            \STATE {Pass signal through physical channel according to \eqref{eq.channel}}
            \STATE {$\mathbf{\hat{r}}_i \leftarrow \mathcal{C}^{-1}(\mathbf{y}_i)$}
            \STATE {$\mathbf{\hat{s}}_i \leftarrow \mathcal{S}^{-1}(\mathbf{\hat{r}}_i, \mathbf{k}_j)$}
            \STATE {$\mathcal{L}_{CE} \leftarrow $ Compute loss function by \eqref{eq.ce}}
            \STATE {Train $\mathcal{{S}(\cdot,\cdot)}$, $\mathcal{S}^{-1}(\cdot,\cdot)$, $\mathcal{{C}(\cdot)}$ and $\mathcal{C}^{-1}(\cdot)$ $\leftarrow$ Gradient descent to minimize $\mathcal{L}_{CE}$}
        \ENDFOR
    \end{algorithmic} 
    \label{alg.train}
\end{algorithm}

\subsection{Channel Encoder and Decoder}
Since this paper mainly focuses on the design of semantic encoder and decoder with the shared knowledge base, the implementations of channel encoder and decoder have not been tailored. For channel encoder, a dense layer and a power normalization layer are utilized to cope with channel noise and normalize the transmit power, respectively. 
In other words, channel encoder maps the residual information $\mathbf{r} \in \mathbb{R}^{L_s\times E}$ into transmitted symbols $\mathbf{x} \in \mathbb{R}^{L_s\times 2M}$, in which $M$ represents the number of transmitted symbols per sentence since DNNs can only process real numbers.
After passing through the channel, a dense layer is used to alleviate the adverse impacts of channel noise on estimated residual information $\hat{\mathbf{r}} \in \mathbb{R}^{L_s\times E}$ from the received symbols $\mathbf{y} \in \mathbb{R}^{L_s\times 2M}$.

\subsection{Semantic Decoder}
Similar to the semantic encoder, the semantic decoder is spilt into two modules, i.e., semantic integration module, which integrates the received residual information and the processed knowledge, and semantic recovery module, which converts symbols back into message.

\subsubsection{Semantic Integration Module}
As shown in Fig. \ref{Fig.structure}, the semantic integration module in semantic decoder has the same network structure as that in semantic encoder, whereas the only difference is the input data. Considering that the residual information obtained contains only part of the source information, the semantic information of the source message can be achieved by fusing the denoised residual information $\hat{\mathbf{r}}$ and the processed knowledge $\mathbf{k'}$. To this end, a multi-head attention layer is adopted and given by
\begin{equation}
    \mathbf{\hat{r}'}_1 = MultiHead (\mathbf{k'},\mathbf{r'},\mathbf{r'}),
\end{equation}
where the inputs are $\mathbf{k'}$ as query, and $\hat{\mathbf{r}}$ as key and value. The output of this module is $\mathbf{\hat{s}'} \in \mathbb{R}^{L_s\times E}$ representing semantic features.

\subsubsection{Semantic Recovery Module}
Since the semantic recovery module, which converts semantic vectors back into messages, can be regarded as the inverse process of the semantic extraction module, the structural design of these two modules are similar. To be specific, the input data $\mathbf{\hat{s}'}$ is sequentially processed by a multi-head attention layer, a normalization layer with residual connection, a dense layer and another normalization layer with residual connection.
Afterwards, in order to transform the data into messages, the data with shape of $\mathbb{R}^{L_s\times E}$ is mapped to the shape of $\mathbb{R}^{L_s\times N_{vocab}}$ through a dense layer. Meanwhile, a softmax layer containing $Softmax(\cdot)$ function is followed, converting the output values to a probability distribution in the range $[0, 1]$ with a sum of $1$.

\subsection{DNN Training Algorithm}
Based on the constructed knowledge base, the transmitter and the receiver networks, including $\mathcal{{S}(\cdot,\cdot)}$, $\mathcal{S}^{-1}(\cdot,\cdot)$, $\mathcal{{C}(\cdot)}$ and $\mathcal{C}^{-1}(\cdot)$, can be optimized simultaneously as illustrated in Algorithm \ref{alg.train}. During the training stage, for training data $\mathbf{s}_i$ in $\mathfrak{S}$, the corresponding knowledge $\mathbf{k}_j$ in $\mathfrak{K}$ is found and used for subsequent processing. The noise in the channel is generated by a random SNR value ranging from $0$ dB to $10$ dB.

\begin{table}
    \renewcommand{\arraystretch}{1.4}
    \centering
    \caption{Simulation Parameters}
    \begin{tabular}{ll} 
    \toprule
    Parameters                                             & Value           \\ 
    \hline
    Pre-configured threshold, $\theta$                      & 0.3             \\
    Embedding dimension, $E$                                & 128             \\
    Number of subspaces in $MulitiHead(\cdot,\cdot,\cdot)$, $h$ & 8               \\
    Dimension used in $Attention(\cdot,\cdot,\cdot)$, $d$       & 128             \\
    Number of transmitted symbols per word, $M$             & 6               \\
    Number of vocabularies, $N_{vocab}$                       & 30527           \\
    Training SNR range                                     & 0 dB $\sim$ 10 dB    \\
    Learning rate                                          & $1\times10^{-2}$  \\
    Batch size                                             & 256             \\
    Training epochs                                        & 100             \\
    \bottomrule
    \label{tab.simulation}
    \end{tabular}
\end{table}

\section{Performance Evaluation}
\label{Sec.Performance Evaluation}
In this section, we will evaluate the performance of the proposed semantic communication system with benchmarks. To give more insight, simulations are carried out under the randomly generated AWGN, Rayleigh and Rician (K factor $=~10~dB$) channel, where the accurate CSI is assumed. Moreover, in order to verify the effectiveness of the knowledge base, knowledge bases constructed based on different threshold $\theta$ are investigated. 

\begin{figure*}[t]
    \centering
    \subfloat[AWGN]{
        \includegraphics[scale=0.65]{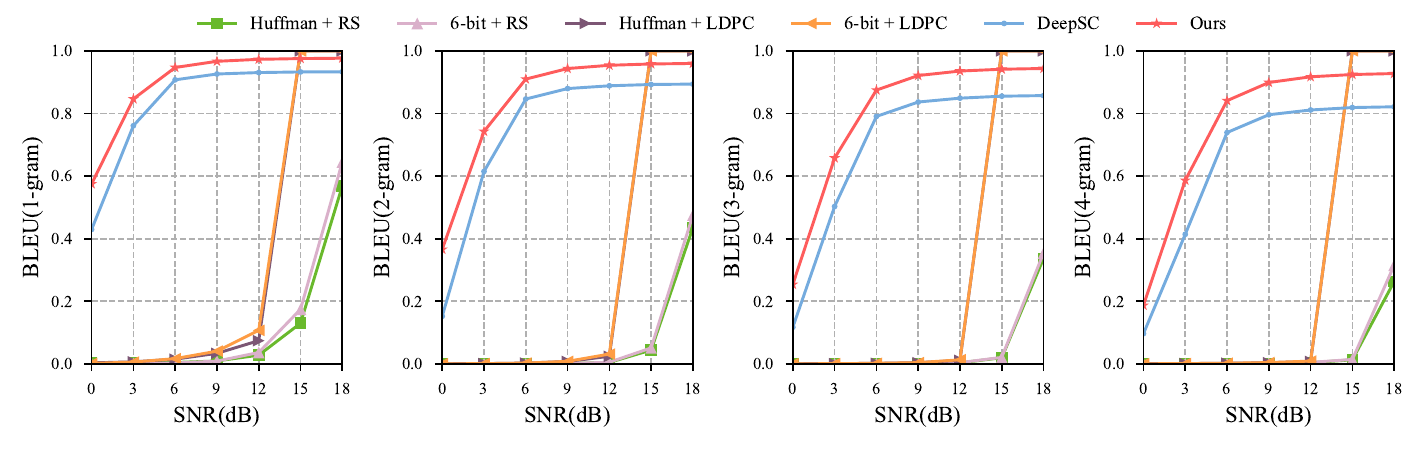}
        }\\
    \subfloat[Rician]{
        \includegraphics[scale=0.65]{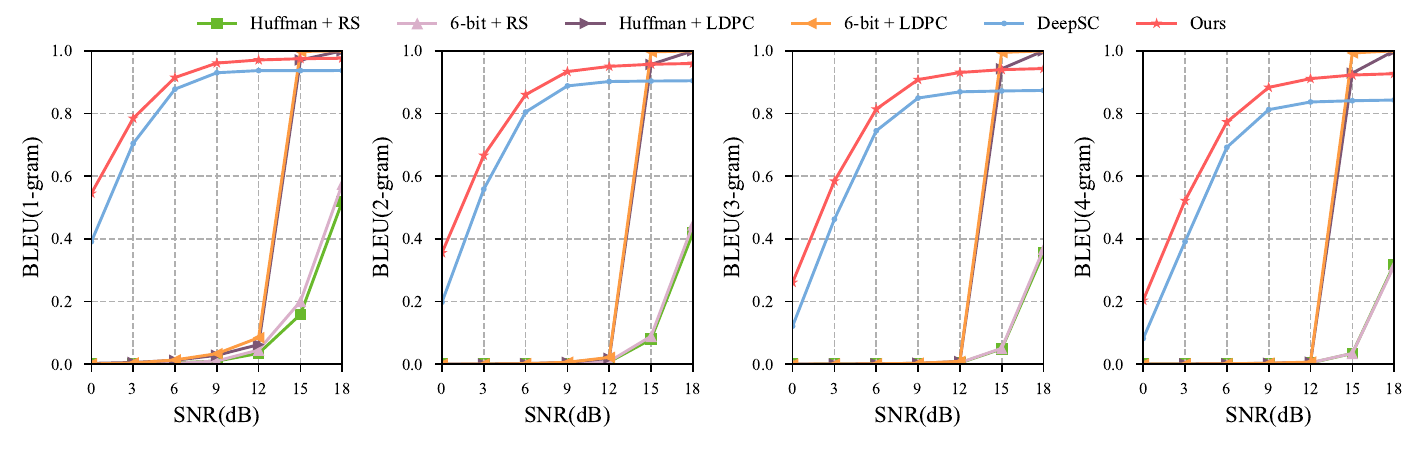}
        }\\
    \subfloat[Rayleigh]{
        \includegraphics[scale=0.65]{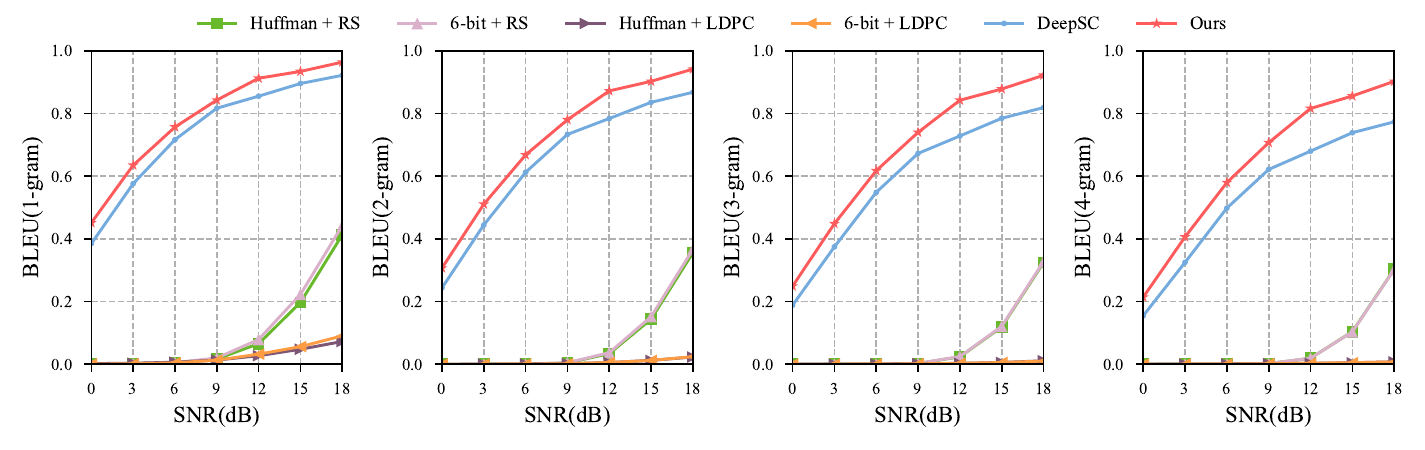}
        }\\
    \caption{BLEU score versus SNR under (a) AWGN channel, (b) Rician channel ($K=10dB$) and (c) Rayleigh channel with Huffman code with RS in 64-QAM; 6-bit code with RS in QAM-64; Huffman code with LDPC in 64-QAM; 6-bit code with LDPC in QAM-64; an end-to-end semantic communication system, named DeepSC \cite{DBLP:journals/tsp/XieQLJ21}, trained over AWGN channel; the proposed system with a shared knowledge base.}
    \label{Fig.BLEU}
    \vspace{-1em}
\end{figure*}

\subsection{Simulation Setting}
\label{Subsec.Simulation Setting}
The European Parliament \cite{DBLP:conf/mtsummit/Koehn05} is preprocessed and adopted as the dataset, which is composed of sentences ranging from $5$ to $20$ words in length. The dataset containing a total of $330k$ sentences is then partitioned into a training dataset, a validation dataset and a testing dataset based on the ratio of $8:1:1$. Unless otherwise specified, the pre-configured threshold $\theta$ of the knowledge base is set as $0.3$.
In our model, the embedding dimension $E$ and the number of transmitted symbols per word $M$ are set to be $128$ and $6$, respectively. Moreover, the $MultiHead$ function we used in our model is initialized with $d = 128$ and $h = 8$. Besides, the Adadelta optimizer with a learning rate of $1\times 10^{-2}$, batch size of $256$, and training epochs of $100$ is adopted in our experiments, as well as a cosine annealing learning rate scheduler \cite{DBLP:conf/iclr/LoshchilovH17}. All simulations are performed by the computer with Intel Core i7-11700K @ 3.60GHz and NVIDIA RTX 3090. Overall, the simulation parameters are illustrated in Table \ref{tab.simulation}.

To justify the effectiveness of the proposed scheme, typical approaches including DeepSC \cite{DBLP:journals/tsp/XieQLJ21} and traditional separate source-channel coding methods are adopted as benchmarks. Specifically, DeepSC is an end-to-end semantic communication system based on joint source-channel coding \cite{DBLP:conf/icassp/FarsadRG18}, which performs source coding and channel coding simultaneously, converting the source message directly into symbols. Moreover, for traditional approaches, Huffman code and fixed-length code (6-bit) are adopted as source coding, and Reed-Solomon (RS) code (5,7) and quasi-cyclic LDPC code are utilized as channel coding. Note that our adopted LDPC code utilizes belief propagation decoding with rate $2/3$, block size $4224$ and Graph $1$ \cite{3gpp.38.212}. 
Besides, in our simulation, the number of distinct character types required to be encoded is $64$, which equals $2^6$. Hence, utilizing additional bits for encoding in fixed-length code would result in significant and unnecessary redundancy. 
Furthermore, Quadrature Amplitude Modulation (QAM) $64$ is utilized for modulation in all traditional approaches. 
Additionally, the index of the knowledge base in our method is assumed to be transmitted error-free through control plane \cite{3GPP} with polar codes \cite{DBLP:journals/comsur/EgilmezXMH20} as channel coding method, and quadrature phase shift keying (QPSK) as modulation method. 

\begin{figure*}[t]
    \centering
    \subfloat[\small{AWGN}]{
        \includegraphics[scale=0.45]{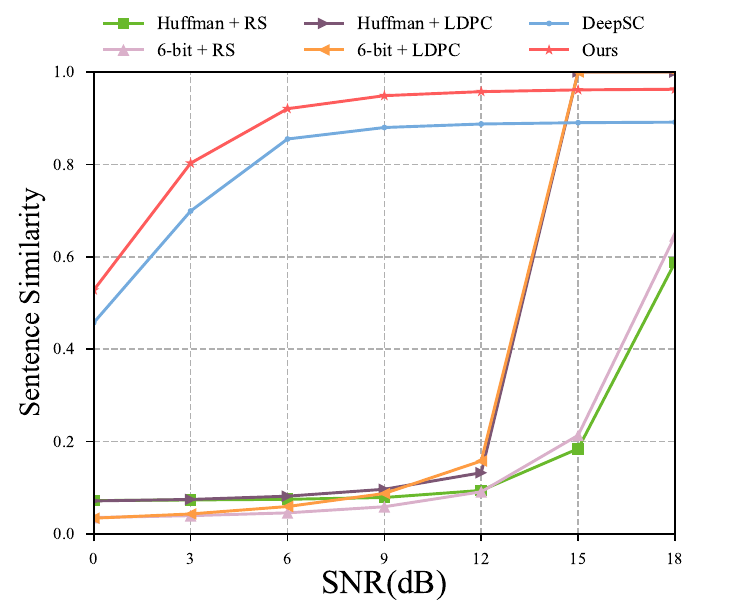}
        }
    \subfloat[\small{Rician}]{
        \includegraphics[scale=0.45]{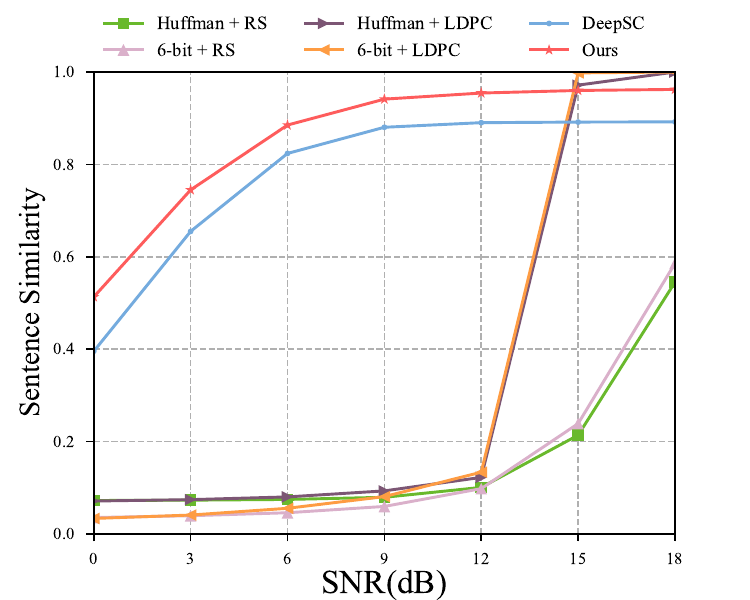}
        }
    \subfloat[\small{Rayleigh}]{
        \includegraphics[scale=0.45]{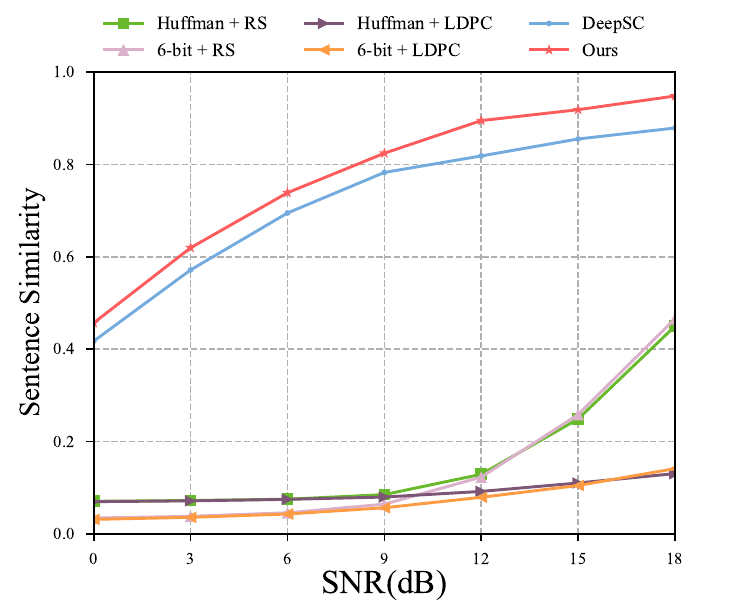}
        }
    \caption{Sentence similarity versus SNR under (a) AWGN channel, (b) Rician channel ($K=10dB$) and (c) Rayleigh channel with Huffman code with RS in 64-QAM; 6-bit code with RS in QAM-64; Huffman code with LDPC in 64-QAM; 6-bit code with LDPC in QAM-64; an end-to-end semantic communication system, named DeepSC \cite{DBLP:journals/tsp/XieQLJ21}; the proposed system with a shared knowledge base.}
    \label{Fig.SS}
\end{figure*}

\subsection{Results and Analysis}
Fig. \ref{Fig.BLEU} shows the relationship between the BLEU score versus SNR under AWGN, Rician and Rayleigh channels. 
It can be observed that conventional methods fail in transmission at low SNR, which may be due to the fact that the conventional methods encode each character, while the DL-enabled methods encode words. To be specific, when some errors occur, character distortions are distributed over many words, and a character error leads to an error in the entire word, resulting in a low BLEU score in traditional approaches. From another point of view, the received word is correct only if all characters in the word are transmitted correctly. 
Moreover, when subjected to high-SNR conditions, LDPC demonstrates excellent performance in both AWGN and Rician channels, capitalizing on its soft decision decoding capabilities. On the other hand, its performance in Rayleigh channel is relatively poor as a result of inaccurate estimation of noise variance. 
Compared with DeepSC, our approach can achieve a higher BLEU score both in low and high SNR regimes. This is because DeepSC only captures the dominant semantic features of sentences, while a few different semantic features among similar sentences are ignored, resulting in inaccurate message recovery. While in our system, only features not in the knowledge base are extracted, which means identical features between similar sentences do not need to be transmitted and unique features can be figured out. For example, when a sentence, ``I'd like to have a sandwich,'' is transmitted and the corresponding knowledge is ``I want something to eat,'' the residual information may represent the word ``sandwich,'' which highly compresses the source sentence with the help of the knowledge base. At the receiver side, the residual information is fused with the same knowledge to recover the original information. In summary, under AWGN, Rician and Rayleigh channels, $12.87\%$, $11.71\%$ and $10.76\%$ improvement compared to DeepSC can be obtain in our system.

Fig. \ref{Fig.SS} shows the sentence similarity score with respect to the SNR under AWGN, Rician and Rayleigh channels. 
It can be observed that our method is competitive to DeepSC and maintains a certain degree of advantage in different channel conditions. Specifically, compared to DeepSC, our system can achieve $9.37\%$, $9.81\%$ and $7.61\%$ improvement on AWGN, Rician and Rayleigh channels, respectively. Compared with the BLEU score, the semantic similarity metric improves marginally. This is due to the fact that errors in a few unimportant words have little influence on human comprehension when the majority of the words are correct.

\begin{table*}
    \renewcommand{\arraystretch}{1.4}
    \centering
    \caption{The average number of symbols sent per sentence and model parameters.}
    \label{tab:my_label}
    \begin{tabular}{ccccccc} 
    \toprule
    \multicolumn{1}{c}{}      & \multicolumn{6}{c}{Methods}             \\ 
    \cline{2-7}
    \vspace{2pt}
                               & Huffman+RS & 6-bit+RS & Huffman + LDPC & 6-bit + LDPC & DeepSC & Ours   \\ 
    \hline
    \vspace{2pt}
    Symbols per sentence       & 80.03      & 112.78 & 89.09 & 119.11 & 131.58 & 106.68  \\ 
    \hline
    \vspace{2pt}
    Parameters ($\times 10^6$) & -     & -   & - & -    & 18.15  & 18.17  \\
    \bottomrule
    \end{tabular}
\end{table*}

Table \ref{tab:my_label} illustrates the average number of symbols sent per sentence, which can reflect the communication efficiency. Benefiting from the advantages of variable-length code, Huffman code requires a relatively small number of symbols per sentence for transmission on average compared to fixed-length code. However, both Huffman code and fixed-length code fail in transmission when SNR is low, which means comparing traditional approaches with DL-enabled approaches in this metric (i.e., symbols per sentence) is meaningless. As for DL-enabled approaches, the number of symbols that need to be transmitted per word is fixed by the network structure. Through rational design of network structure, our approach requires less symbols per sentence compared with DeepSC while achieves better performance. This is due to the introduction of the knowledge base that allows the system to communicate with only a small amount of residual information instead of the whole semantic information. Table \ref{tab:my_label} also illustrates the number of parameters for DL-enabled model, which indicates the complexity of the model and the required storage size. 
In general, compared with DeepSC, our approach can achieve $11.77\%$ and $8.93\%$ average improvement on BLEU score and sentence similarity score under AWGN channel, respectively, with $81\%$ transmitted symbols and a similar level of the number of parameters. 
It is noteworthy that the performance gap between our proposed method and DeepSC is derived by the shared knowledge base, since they are based on the same network structure (i.e., Transformer model \cite{DBLP:conf/nips/VaswaniSPUJGKP17}) and have a similar number of parameters. 

\begin{figure}[t]
    \setlength{\abovecaptionskip}{-0.1cm}
    \setlength{\belowcaptionskip}{-1.5cm}
    \centering
    \includegraphics[width = 0.45\textwidth]{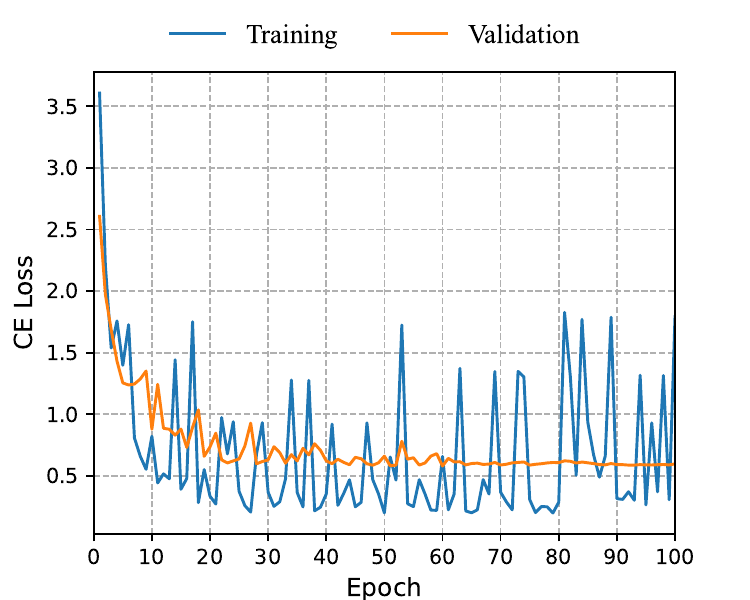}
    \caption{The CE loss versus epoch for training with SNR ranging from $0$ dB to $10$ dB and validation with SNR = $3$ dB under the AWGN channel.}
    \label{Fig.Loss}
\end{figure}

To demonstrate the convergence of our model, Fig. \ref{Fig.Loss} draws the relationship between the CE loss value \eqref{eq.ce} and the increasing epoch under training and validation mode. 
At the early stage of training, both training loss and validation loss decrease rapidly. As mentioned, our model is trained with a random SNR ranging from $0$ dB to $10$ dB, which explains the large fluctuations in training loss at the later stage of training. As shown in Fig. \ref{Fig.Loss}, the loss value under validation mode with fixed SNR = $3$ dB gradually converges to a stable state, which proves that our model has converged. 

\begin{figure}[t]
    \setlength{\abovecaptionskip}{-0.1cm}
    \setlength{\belowcaptionskip}{-1.5cm}
    \centering
    \includegraphics[width = 0.45\textwidth]{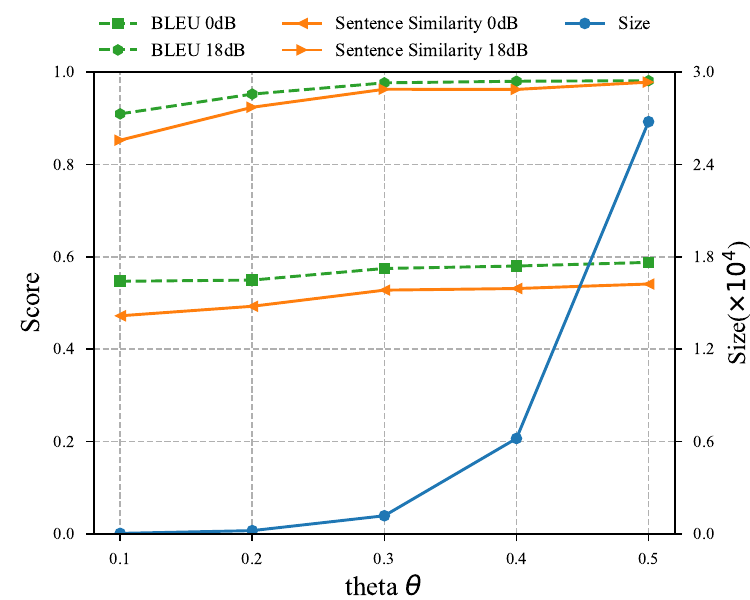}
    \caption{The different pre-configured thresholds $\theta$ versus the performance and the size of the knowledge base under the AWGN channel. Specifically, BLEU (1-gram) score and sentence similarity score are adopted as performance metrics, and the size refers to the number of sentences contained in the knowledge.}
    \label{Fig.knowledge}
\end{figure}

To investigate the critical role of the shared knowledge base, a comparative experiment is employed, in which different thresholds $\theta$ (i.e., $0.1$, $0.2$, $0.3$, $0.4$, and $0.5$) are used to construct our knowledge bases. On the basis of these knowledge bases, the networks are trained and evaluated under AWGN channel, as shown in Fig. \ref{Fig.knowledge}. 
It can be observed that the performance metrics, including BLEU (1-gram) score and sentence similarity score, improve with the increase of the threshold $\theta$, i.e., the size of the knowledge base, which proves the effectiveness of the shared knowledge base. Generally, when the size of the knowledge base grows, the amount of residual information that needs to be sent will decrease. 
However, the improvement in performance is not very significant, since the number of commonly used semantic messages is limited. Besides, the size of the knowledge base (i.e., the number of sentences contained in the knowledge) will grow dramatically with the increase of threshold $\theta$. Therefore, it is essential to have an appropriate threshold value selected.

\section{Conclusions}
\label{Sec.Conclusions}
In this paper, a novel semantic communication system with a shared knowledge base is proposed for text transmission. With the aid of the shared knowledge base, the proposed system integrates the message and corresponding knowledge from the shared knowledge base to obtain the residual information, which enables the system to transmit fewer symbols without semantic performance degradation. 
To make our system more reliable, the semantic self-information of each message and the entropy of the source are mathematically defined based on the knowledge base. 
Furthermore, the knowledge base construction algorithm is developed based on a similarity-comparison method, in which a pre-configured threshold can be leveraged to control the size of the knowledge base. 
Moreover, the simulation results have demonstrated that the proposed system has a definite advantage over conventional communication systems in terms of BLEU and sentence similarity metrics, especially in the case of low SNR regime. Compared to DeepSC, our system can achieve $11.77\%$ and $8.93\%$ average improvement on BLEU score and sentence similarity score over AWGN channel, respectively. At the same time, the average number of symbols transmitted per sentence has been reduced by $19\%$. 

The complexity of human language necessitates the evolution of semantic encoders and decoders into large language models, leading to significant storage requirements and longer reasoning time. In this paper, the proposed system offers a potential solution to alleviate these challenges, which can be widely used for text and image transmission. 
Moving forward, there exist numerous fascinating research directions that warrant further exploration, such as how to design comprehensive update and synchronization mechanisms for the knowledge base, how to design personalized knowledge bases for different users, and how to extend the text knowledge base into a multi-modal one.

\bibliographystyle{IEEEtran}
\bibliography{References}

\begin{thebibliography}{10}
\providecommand{\url}[1]{#1}
\csname url@samestyle\endcsname
\providecommand{\newblock}{\relax}
\providecommand{\bibinfo}[2]{#2}
\providecommand{\BIBentrySTDinterwordspacing}{\spaceskip=0pt\relax}
\providecommand{\BIBentryALTinterwordstretchfactor}{4}
\providecommand{\BIBentryALTinterwordspacing}{\spaceskip=\fontdimen2\font plus
\BIBentryALTinterwordstretchfactor\fontdimen3\font minus
  \fontdimen4\font\relax}
\providecommand{\BIBforeignlanguage}[2]{{%
\expandafter\ifx\csname l@#1\endcsname\relax
\typeout{** WARNING: IEEEtran.bst: No hyphenation pattern has been}%
\typeout{** loaded for the language `#1'. Using the pattern for}%
\typeout{** the default language instead.}%
\else
\language=\csname l@#1\endcsname
\fi
#2}}
\providecommand{\BIBdecl}{\relax}
\BIBdecl

\bibitem{DBLP:journals/bstj/Shannon48}
C.~E. Shannon, ``A mathematical theory of communication,'' \emph{Bell Syst.
  Tech. J.}, vol.~27, no.~3, pp. 379--423, 1948.

\bibitem{DBLP:journals/comsur/FangBGL15}
Y.~Fang, G.~Bi, Y.~L. Guan, and F.~C.~M. Lau, ``A survey on protograph {LDPC}
  codes and their applications,'' \emph{{IEEE} Commun. Surv. Tutorials},
  vol.~17, no.~4, pp. 1989--2016, 2015.

\bibitem{DBLP:journals/comsur/EgilmezXMH20}
Z.~B.~K. Egilmez, L.~Xiang, R.~G. Maunder, and L.~Hanzo, ``The development,
  operation and performance of the {5G} polar codes,'' \emph{{IEEE} Commun.
  Surv. Tutorials}, vol.~22, no.~1, pp. 96--122, 2020.

\bibitem{DBLP:journals/cm/KaiNiu22}
K.~Niu, J.~Dai, S.~Yao, S.~Wang, Z.~Si, X.~Qin, and P.~Zhang, ``A paradigm
  shift toward semantic communications,'' \emph{{IEEE} Commun. Mag.}, vol.~60,
  no.~11, pp. 113--119, 2022.

\bibitem{SIX-Trust_KangXin}
Y.~Wang, X.~Kang, T.~Li, H.~Wang, C.~Chu, and Z.~Lei, ``{SIX-Trust} for {6G}:
  Toward a secure and trustworthy future network,'' \emph{{IEEE} Access},
  vol.~11, pp. 107\,657--107\,668, 2023.

\bibitem{DBLP:journals/wc/ChenLSKCP20}
S.~Chen, Y.-C. Liang, S.~Sun, S.~Kang, W.~Cheng, and M.~Peng, ``Vision,
  requirements, and technology trend of {6G}: How to tackle the challenges of
  system coverage, capacity, user data-rate and movement speed,'' \emph{{IEEE}
  Wirel. Commun.}, vol.~27, no.~2, pp. 218--228, 2020.

\bibitem{DBLP:journals/network/GiordaniZ21}
M.~Giordani and M.~Zorzi, ``Non-terrestrial networks in the {6G} era:
  Challenges and opportunities,'' \emph{{IEEE} Netw.}, vol.~35, no.~2, pp.
  244--251, 2021.

\bibitem{zhang2022toward}
P.~Zhang, W.~Xu, H.~Gao, K.~Niu, X.~Xu, X.~Qin, C.~Yuan, Z.~Qin, H.~Zhao,
  J.~Wei \emph{et~al.}, ``Toward wisdom-evolutionary and primitive-concise
  {6G}: A new paradigm of semantic communication networks,''
  \emph{Engineering}, vol.~8, pp. 60--73, 2022.

\bibitem{wheeler2022engineering}
D.~Wheeler and B.~Natarajan, ``Engineering semantic communication: A survey,''
  \emph{IEEE Access}, vol.~11, pp. 13\,965--13\,995, 2023.

\bibitem{DBLP:conf/globecom/HuangTGL21}
D.~Huang, X.~Tao, F.~Gao, and J.~Lu, ``Deep learning-based image semantic
  coding for semantic communications,'' in \emph{{IEEE} Global Communications
  Conference, {GLOBECOM} 2021, Madrid, Spain, December 7-11, 2021}.\hskip 1em
  plus 0.5em minus 0.4em\relax {IEEE}, 2021.

\bibitem{DBLP:journals/wc/LuoCG22}
X.~Luo, H.~Chen, and Q.~Guo, ``Semantic communications: Overview, open issues,
  and future research directions,'' \emph{{IEEE} Wirel. Commun.}, vol.~29,
  no.~1, pp. 210--219, 2022.

\bibitem{DBLP:journals/corr/abs-2201-01389}
\BIBentryALTinterwordspacing
Z.~Qin, X.~Tao, J.~Lu, and G.~Y. Li, ``Semantic communications: Principles and
  challenges,'' \emph{CoRR}, vol. abs/2201.01389, 2022. [Online]. Available:
  \url{https://arxiv.org/abs/2201.01389}
\BIBentrySTDinterwordspacing

\bibitem{DBLP:books/aw/RN2020}
\BIBentryALTinterwordspacing
S.~Russell and P.~Norvig, \emph{Artificial Intelligence: {A} Modern Approach
  (4th Edition)}.\hskip 1em plus 0.5em minus 0.4em\relax Pearson, 2020.
  [Online]. Available: \url{http://aima.cs.berkeley.edu/}
\BIBentrySTDinterwordspacing

\bibitem{DBLP:journals/tccn/OSheaH17}
T.~J. O'Shea and J.~Hoydis, ``An introduction to deep learning for the physical
  layer,'' \emph{{IEEE} Trans. Cogn. Commun. Netw.}, vol.~3, no.~4, pp.
  563--575, 2017.

\bibitem{DBLP:journals/jcin/LanWZZCPH21}
Q.~Lan, D.~Wen, Z.~Zhang, Q.~Zeng, X.~Chen, P.~Popovski, and K.~Huang, ``What
  is semantic communication? {A} view on conveying meaning in the era of
  machine intelligence,'' \emph{J. Commun. Inf. Networks}, vol.~6, no.~4, pp.
  336--371, 2021.

\bibitem{DBLP:journals/cn/StrinatiB21}
E.~C. Strinati and S.~Barbarossa, ``{6G} networks: Beyond shannon towards
  semantic and goal-oriented communications,'' \emph{Comput. Networks}, vol.
  190, p. 107930, 2021.

\bibitem{DBLP:conf/icc/HanY0H022}
T.~Han, Q.~Yang, Z.~Shi, S.~He, and Z.~Zhang, ``Semantic-preserved
  communication system for highly efficient speech transmission,'' \emph{{IEEE}
  J. Sel. Areas Commun.}, vol.~41, no.~1, pp. 245--259, 2023.

\bibitem{DBLP:journals/tcom/KangSGQY22}
X.~Kang, B.~Song, J.~Guo, Z.~Qin, and F.~R. Yu, ``Task-oriented image
  transmission for scene classification in unmanned aerial systems,''
  \emph{{IEEE} Trans. Commun.}, vol.~70, no.~8, pp. 5181--5192, 2022.

\bibitem{DBLP:conf/mlsp/YiCXL22}
P.~Yi, Y.~Cao, J.~Xu, and Y.-C. Liang, ``Semantic communication for remote
  spectrum sensing in non-terrestrial networks,'' in \emph{32nd {IEEE}
  International Workshop on Machine Learning for Signal Processing, {MLSP},
  Xi'an, China, August 22-25, 2022}.\hskip 1em plus 0.5em minus 0.4em\relax
  {IEEE}, 2022.

\bibitem{DBLP:journals/jsac/XieQTL22}
H.~Xie, Z.~Qin, X.~Tao, and K.~B. Letaief, ``Task-oriented multi-user semantic
  communications,'' \emph{{IEEE} J. Sel. Areas Commun.}, vol.~40, no.~9, pp.
  2584--2597, 2022.

\bibitem{DBLP:journals/icl/HuZZWHZ22}
H.~Hu, X.~Zhu, F.~Zhou, W.~Wu, R.~Q. Hu, and H.~Zhu, ``One-to-many semantic
  communication systems: Design, implementation, performance evaluation,''
  \emph{{IEEE} Commun. Lett.}, vol.~26, no.~12, pp. 2959--2963, 2022.

\bibitem{DBLP:journals/tsp/XieQLJ21}
H.~Xie, Z.~Qin, G.~Y. Li, and B.~Juang, ``Deep learning enabled semantic
  communication systems,'' \emph{{IEEE} Trans. Signal Process.}, vol.~69, pp.
  2663--2675, 2021.

\bibitem{DBLP:journals/jsac/WengQ21}
Z.~Weng and Z.~Qin, ``Semantic communication systems for speech transmission,''
  \emph{{IEEE} J. Sel. Areas Commun.}, vol.~39, no.~8, pp. 2434--2444, 2021.

\bibitem{DBLP:journals/jsac/XieQ21}
H.~Xie and Z.~Qin, ``A lite distributed semantic communication system for
  internet of things,'' \emph{{IEEE} J. Sel. Areas Commun.}, vol.~39, no.~1,
  pp. 142--153, 2021.

\bibitem{DBLP:journals/jsac/WangCLSNPC22}
Y.~Wang, M.~Chen, T.~Luo, W.~Saad, D.~Niyato, H.~V. Poor, and S.~Cui,
  ``Performance optimization for semantic communications: An attention-based
  reinforcement learning approach,'' \emph{{IEEE} J. Sel. Areas Commun.},
  vol.~40, no.~9, pp. 2598--2613, 2022.

\bibitem{DBLP:journals/tcom/JiangWJL22}
P.~Jiang, C.~Wen, S.~Jin, and G.~Y. Li, ``Deep source-channel coding for
  sentence semantic transmission with {HARQ},'' \emph{{IEEE} Trans. Commun.},
  vol.~70, no.~8, pp. 5225--5240, 2022.

\bibitem{DBLP:journals/corr/abs-2112-03093}
J.~Dai, P.~Zhang, K.~Niu, S.~Wang, Z.~Si, and X.~Qin, ``Communication beyond
  transmitting bits: Semantics-guided source and channel coding,'' \emph{{IEEE}
  Wirel. Commun.}, vol.~30, no.~4, pp. 170--177, 2023.

\bibitem{yao2022semantic}
S.~Yao, K.~Niu, S.~Wang, and J.~Dai, ``Semantic coding for text transmission:
  An iterative design,'' \emph{{IEEE} Trans. Cogn. Commun. Netw.}, pp. 1--1,
  2022.

\bibitem{ICC2023}
P.~Yi, Y.~Cao, X.~Kang, and Y.-C. Liang, ``Semantic communication systems with
  a shared knowledge base,'' in \emph{2023 IEEE International Conference on
  Communications Workshops (ICC Workshops)}, 2023, pp. 1374--1379.

\bibitem{ribeiro2016should}
M.~T. Ribeiro, S.~Singh, and C.~Guestrin, ``'why should {I} trust you?'
  explaining the predictions of any classifier,'' in \emph{Proceedings of the
  22nd ACM SIGKDD international conference on knowledge discovery and data
  mining}, 2016, pp. 1135--1144.

\bibitem{bao2011towards}
J.~Bao, P.~Basu, M.~Dean, C.~Partridge, A.~Swami, W.~Leland, and J.~A. Hendler,
  ``Towards a theory of semantic communication,'' in \emph{2011 IEEE Network
  Science Workshop}, 2011, pp. 110--117.

\bibitem{DBLP:journals/iandc/LucaT74}
A.~de~Luca and S.~Termini, ``Entropy of l-fuzzy sets,'' \emph{Inf. Control.},
  vol.~24, no.~1, pp. 55--73, 1974.

\bibitem{wang2017distributed}
S.~Wang, Y.~Fang, and S.~Cheng, \emph{Distributed Source Coding: Theory and
  Practice}.\hskip 1em plus 0.5em minus 0.4em\relax John Wiley \& Sons, 2017.

\bibitem{DBLP:books/daglib/0091821}
T.~S. Rappaport, \emph{Wireless communications - principles and
  practice}.\hskip 1em plus 0.5em minus 0.4em\relax Prentice Hall, 1996.

\bibitem{3GPP}
``System architecture for the {5G} system,'' {3GPP}, TS 23.501, 03 2022,
  {Version} 15.13.0.

\bibitem{DBLP:conf/acl/PapineniRWZ02}
K.~Papineni, S.~Roukos, T.~Ward, and W.~Zhu, ``Bleu: a method for automatic
  evaluation of machine translation,'' in \emph{Proceedings of the 40th Annual
  Meeting of the Association for Computational Linguistics, July 6-12, 2002,
  Philadelphia, PA, {USA}}.\hskip 1em plus 0.5em minus 0.4em\relax {ACL}, 2002,
  pp. 311--318.

\bibitem{DBLP:conf/emnlp/ReimersG19}
N.~Reimers and I.~Gurevych, ``Sentence-{BERT}: Sentence embeddings using
  siamese {BERT}-networks,'' in \emph{Proceedings of the 2019 Conference on
  Empirical Methods in Natural Language Processing and the 9th International
  Joint Conference on Natural Language Processing (EMNLP-IJCNLP)}, 2019, pp.
  3982--3992.

\bibitem{DBLP:journals/cm/ShiXLX21}
G.~Shi, Y.~Xiao, Y.~Li, and X.~Xie, ``From semantic communication to
  semantic-aware networking: Model, architecture, and open problems,''
  \emph{{IEEE} Commun. Mag.}, vol.~59, no.~8, pp. 44--50, 2021.

\bibitem{jo2008single}
T.-H. Jo, ``Single pass algorithm for text clustering by encoding documents
  into tables,'' \emph{Journal of Korea Multimedia Society}, vol.~11, no.~12,
  pp. 1749--1757, 2008.

\bibitem{DBLP:conf/nips/VaswaniSPUJGKP17}
A.~Vaswani, N.~Shazeer, N.~Parmar, J.~Uszkoreit, L.~Jones, A.~N. Gomez,
  L.~Kaiser, and I.~Polosukhin, ``Attention is all you need,'' in
  \emph{Advances in Neural Information Processing Systems 30: Annual Conference
  on Neural Information Processing Systems 2017, December 4-9, 2017, Long
  Beach, CA, {USA}}, 2017, pp. 5998--6008.

\bibitem{DBLP:journals/corr/BaKH16}
\BIBentryALTinterwordspacing
L.~J. Ba, J.~R. Kiros, and G.~E. Hinton, ``Layer normalization,'' \emph{CoRR},
  vol. abs/1607.06450, 2016. [Online]. Available:
  \url{http://arxiv.org/abs/1607.06450}
\BIBentrySTDinterwordspacing

\bibitem{DBLP:journals/corr/HendrycksG16}
\BIBentryALTinterwordspacing
D.~Hendrycks and K.~Gimpel, ``Bridging nonlinearities and stochastic
  regularizers with gaussian error linear units,'' \emph{CoRR}, vol.
  abs/1606.08415, 2016. [Online]. Available:
  \url{http://arxiv.org/abs/1606.08415}
\BIBentrySTDinterwordspacing

\bibitem{DBLP:conf/mtsummit/Koehn05}
P.~Koehn, ``Europarl: {A} parallel corpus for statistical machine
  translation,'' in \emph{Proceedings of Machine Translation Summit {X:}
  Papers, MTSummit 2005, Phuket, Thailand, September 13-15, 2005}, 2005, pp.
  79--86.

\bibitem{DBLP:conf/iclr/LoshchilovH17}
I.~Loshchilov and F.~Hutter, ``{SGDR:} stochastic gradient descent with warm
  restarts,'' in \emph{5th International Conference on Learning
  Representations, {ICLR} 2017, Toulon, France, April 24-26, 2017, Conference
  Track Proceedings}.\hskip 1em plus 0.5em minus 0.4em\relax OpenReview.net,
  2017.

\bibitem{DBLP:conf/icassp/FarsadRG18}
N.~Farsad, M.~Rao, and A.~Goldsmith, ``Deep learning for joint source-channel
  coding of text,'' in \emph{2018 {IEEE} International Conference on Acoustics,
  Speech and Signal Processing, {ICASSP} 2018, Calgary, AB, Canada, April
  15-20, 2018}.\hskip 1em plus 0.5em minus 0.4em\relax {IEEE}, 2018, pp.
  2326--2330.

\bibitem{3gpp.38.212}
``{NR}; multiplexing and channel coding,'' {3GPP}, TS 38.212, 07 2018,
  {Version} 15.2.0.

\end{thebibliography}

\vfill

\end{document}